%% file: main.tex
\documentclass[10pt]{article}
\usepackage[preprint]{tmlr}   
\usepackage[hyperfootnotes=false]{hyperref}
\hypersetup{colorlinks=true,linkcolor=blue!50!black,citecolor=blue!50!black,urlcolor=blue!50!black}
\usepackage{url}
\usepackage{graphicx}
\usepackage{booktabs}
\usepackage{amsmath}
\usepackage{listings}
\usepackage{tikz}
\usetikzlibrary{positioning,arrows.meta}
\newcommand{\SV}{\textsc{sv}}

\title{Selecting LLM-Generated Hypotheses Raises Truth Beyond What an LLM Panel Expects:\\A Pre-registered Study on NAS-Bench-201 and HPOBench}
\author{\name Shanglin Yang$^{*}$ \email kudoysl@gmail.com\\
      \addr Independent Researcher, USA
      \AND
      \name Zhan Shi$^{*}$ \email ashi2@scu.edu\\
      \addr Santa Clara University, USA}
\def\month{MM}
\def\year{YYYY}
\def\openreview{\url{https://openreview.net/forum?id=XXXX}}

\begin{document}
\maketitle
{\renewcommand{\thefootnote}{*}\footnotetext{Equal contribution.}}

\begin{abstract}
LLM research-ideation systems aim for hypotheses that are both true and non-obvious, yet neither property can normally be checked. We build a testbed in which both can. Generators propose conditional interaction claims about two fully tabulated benchmarks, NAS-Bench-201 and HPOBench; every claim is verified exactly against the tables, and its non-obviousness is scored against a calibrated blind prior elicited from other LLMs. Excess truth measures how much more often a run's claims hold than this panel expects. Across a series of pre-registered stages with Claude and GPT generators, our main positive result is a simple two-stage step found by a harness optimizer: draft a large pool of claims, then select a subset in a separate call. For Sonnet on NAS-Bench-201, this step raised excess truth in every pre-registered test, including against a same-pool ablation that keeps only likely-true claims, without lowering precision. The gain comes both from credit for true claims and from avoiding near-miss false ones. In post hoc checks it held under every calibration, panel and truth rule we tried, and it exceeded random and type-filter selection from the same pools. Because kept claims recur across runs, the evidence weakens when runs and claims are resampled jointly, and it is not significant against Sonnet's own prior. Two secondary findings are also pre-registered. GPT makes real use of the results of cheap probes, beating shuffled and withheld controls, though not more efficiently than asking for more claims. Optimizing the harness for an LLM judge raised judge scores while sharply cutting precision for every generator.
\end{abstract}

\section{Introduction}
\label{sec:intro}
Automated research systems now propose ideas, run experiments and write papers \citep{lu2024scientist,yamada2025scientist}. A growing line of work improves them by optimizing the scaffold around the model: prompts \citep{agrawal2025gepa}, agent workflows \citep{hu2024automated}, and whole harnesses \citep{xu2026harness}. With an executable score this works. Recent systems improve their own research agents by accepting only rewrites that raise a private held-out score \citep{srikanth2026recursive}, or turn papers into executable environments for repeated improvement \citep{wang2026rsiforge}. Research ideation has no such score. Whether an idea is true is known only after someone tests it, and whether it is surprising depends on what the field already believes. Ideation systems are therefore selected with an LLM judge's novelty and quality ratings, or with similarity to later papers \citep{liu2026ideascientist}, and there are reasons to distrust both. Judge-rated novelty is negatively correlated with how ideas fare in hindsight \citep{jiang2026hindsight}, judges reward style over substance \citep{feuer2025style,ji2026style}, and under optimization proxy rewards are exploited \citep{gao2022scaling,zhou2026more}.

We built a testbed in which both halves of the objective can be measured. Hypotheses are conditional claims about two fully tabulated benchmarks, so every claim is checked exactly against the table. Surprise is scored against blind ratings of every possible claim by a panel of other LLMs, calibrated to truth over the whole claim space: a run scores $R>0$ when its claims hold more often than the calibrated panel predicted. In a pre-registered study with Claude and GPT generators we compared direct generation, which asks for ten hypotheses in one call, with evolved harnesses, with different optimizer fitness functions, and with measurement, a budget of about 100 cheap experiments read before proposing. This paper reports what raised verified surprise and how robust each gain is.

\paragraph{Findings.}
\begin{itemize}
\item \textbf{A two-stage selection step raises excess truth relative to a calibrated LLM panel without losing precision.} Section~\ref{sec:sel} reports this finding for one generator on one benchmark. An optimizer-evolved harness, singled out post hoc among 19 harness evaluations and then confirmed in fresh runs, drafts 30 candidate claims and keeps 10 in a second call. Its rule targets claims the generator believes at $\ge 55\%$ but expects other models to doubt at $<50\%$. For Sonnet on NAS-Bench-201, this step raised $R$ by $1.77$ per run over direct generation, by $1.95$ over a new batch of direct runs, and by $1.96$ over a likely-true selection ablation. Each of these three pre-registered tests gave $p<.001$, although all three share the same 24 harness runs. Precision rose from $.79$ to $.83$. Post hoc, the step also beat random and claim-type selection from the identical pools, by $1.55$ and $0.95$ per run. Separately elicited blind priors do not match the rule's literal thresholds, so the rule's wording describes its design prompt rather than a verified internal mechanism.
\item \textbf{GPT uses informative experimental values, though not more cheaply than pooling.} In pre-registered tests reported in Section~\ref{sec:meas}, GPT did better after reading 100 cheap probes than after reading the same probes with shuffled or withheld values, on both benchmarks. A cheaper single-call baseline of 40 claims matched it on HPOBench, however, and on NAS-Bench-201 measurement pushed precision below the pre-registered bound of $0.8$ times direct generation.
\item \textbf{Optimizing a harness for an LLM judge inflated judge ratings while collapsing precision.} Section~\ref{sec:a} shows that the harness evolved for the judge's novelty-plus-quality score raised judge ratings on both benchmarks and for all three generators, while cutting precision by 46--68\%. This result was pre-registered. Post hoc, the drop holds under every truth definition we tested and, on HPOBench, within claim type.
\end{itemize}
\paragraph{What the gain is.} Two qualifications apply to the first finding throughout. The first is uncertainty. Resampling runs alone gives a 90\% interval of $[1.30, 2.25]$ against direct generation, but the kept claims recur across runs, and resampling propositions as well widens it to $[0.38, 3.30]$. Under this joint resampling the gain is above zero against the Claude and Claude+GPT panels, with $p_\times\leq.024$, but not against Sonnet's own blind prior, where $p_\times$ ranges from $.06$ to $.19$. The second is composition. $R$ is credit for true claims, $\sum_{\text{true}}(1-\pi)$, minus cost for false ones, $\sum_{\text{false}}\pi$. At the main truth threshold of 2 points, credit accounts for $+1.04$ of the $+1.77$ gain and is only marginal with propositions resampled, at $p_\times=.054$; the remaining $+0.73$ is avoided cost on false claims, mostly near misses. At thresholds of 1 and 1.5 points credit is nearly the whole gain, and at 3 points credit and avoided cost are about equal. The defensible reading is that the selected claims are true more often than a calibrated Claude panel expects, partly by avoiding claims that miss the threshold narrowly.

Table~\ref{tab:status} gives the status of each claim. The study also produced negative results, which we report in Section~\ref{sec:diag} as lessons for evaluating ideation: our original surprise metric was biased by direction and manufactured a confirmed measurement gain for Claude generators, and other harness variants did not beat direct generation.

\begin{table}[t]
\centering\footnotesize
\begin{tabular}{p{6.6cm}p{2.0cm}p{5.6cm}}
\toprule
Claim & Evidence & Status \\
\midrule
Stage-3 selection step raises $R$ at equal or higher precision (Sonnet, NB201) & Stages 13--15 & Pre-registered; supported in all three tests \\
\quad robust to calibration, panel and truth definition & App.~\ref{app:sens} & Post hoc; holds under all 14 calibrations, 8 panels and 6 truth definitions ($\tau$ 1--3 pp) \\
\quad robust to resampling the recurring propositions & App.~\ref{app:decomp} & Post hoc; yes under the Claude and Claude+GPT panels ($p_\times\leq.024$); not under Sonnet's own prior ($p_\times=.06$--$.19$) \\
\quad beats random and claim-type selection from its own pools & App.~\ref{app:selection} & Post hoc; yes ($+1.55$, $+0.95$); gain concentrated in about 10 propositions \\
\quad gain is credit for true claims, not only avoided errors & App.~\ref{app:decomp} & Post hoc; credit $+1.04$ of $+1.77$; with propositions resampled $p_\times=.028$--$.054$ \\
\quad a selector given the scoring panel's priors does better & App.~\ref{app:selection} & Post hoc; yes ($+3.42$ against the harness's $+2.87$); the harness has no access to them \\
\quad kept claims follow the generator's own beliefs (the rule's wording) & App.~\ref{app:selection} & Post hoc; not supported (27\% of kept vs 45\% of unkept rated $\geq 55$ by Sonnet's blind prior) \\
GPT: measurement uses informative values (beats shuffled and withheld) & Stage 15 & Pre-registered; supported in 4 of 4 \\
GPT: measurement beats same-cost direct baselines on $R$ & Stages 13--15 & Pre-registered on $p$ alone; supported in 5 of 6 tests (refuted against the 40-claim call on HPO). On NB201 precision is .61--.72 times every same-cost arm's, below the 0.8 bound (post hoc reading); per distinct claim, not significant against pooled + rule on HPO (post hoc). No budget-efficient gain shown \\
GPT: measurement keeps precision $\geq 0.8\times$ direct & Stage 10 & Pre-registered; met on HPO, missed on NB201 (.69$\to$.50) \\
Opus: measurement gains with GPT in the panel (NB201) & Stage 12b & Pre-registered as a robustness test of a predicted null, which failed (Holm $p<.05$); panel- and calibration-dependent \\
Opus and GPT gain under informed panels and own prior & Section~\ref{sec:n} & Post hoc; hypothesis-generating (panel quality and frontier-model membership confounded) \\
The one judge-optimized harness: judge score up, precision down & Stages 5, 9, 10 & Pre-registered; supported in every pair (one harness, one optimizer run) \\
\quad precision lower within claim type & App.~\ref{app:checks} & Post hoc; yes on HPO; cannot be separated from type shift on NB201 \\
Measurement raises \SV{} for Claude generators & Stages 5, 6, 9 & Supported under the original metric; refuted by the pre-registered stage-12 rescoring (Section~\ref{sec:s}) \\
\bottomrule
\end{tabular}
\caption{Status of the paper's claims. Table~\ref{tab:ledger} lists every pre-registered criterion and its verdict. $R$ was introduced post hoc in stage 12 and is pre-registered from stage 12b on.}
\label{tab:status}
\end{table}

\section{Related Work}
\paragraph{Ideation systems and their evaluation.}
Systems generate ideas with retrieval, role pipelines, evolutionary search or role-wise RL \citep{baek2024researchagent,li2024chain,liu2026ideascientist}, and some rank hypotheses in tournaments judged by LLMs \citep{gottweis2025accelerating}. They are evaluated by human experts \citep{si2024llms}, by executing the ideas \citep{si2025ideation}, against datasets that can verify a hypothesis \citep{majumder2024discoverybench,liu2025hypobench}, or by LLM judges, whose reliability has been studied statically \citep{jiang2026hindsight,ji2026style,sinhahajari2026limits}; domain experts disagree with judges that call model-written research questions highly novel \citep{sinhahajari2026limits}. Simulated environments test the full discovery cycle \citep{jansen2024discoveryworld}. Audits of AI-scientist systems find failures that final outputs hide \citep{luo2025more}. In science itself, surprising combinations of contents and contexts are associated with impact \citep{shi2023surprising}. Outcome predictors have been proposed as rewards \citep{wen2025predicting}. We check ideas against measured truth inside an optimization loop and score surprise against a calibrated panel.

\paragraph{Eliciting probabilities and surprise.}
Forecasting the outcomes of experiments has been studied for people and models \citep{dellavigna2019predict,luo2024brainbench,hewitt2024predicting,wen2025predicting}. Judged probabilities depend on how the outcome space is described \citep{tversky1994support} and are pulled toward a uniform split over whatever partition is offered \citep{fox2005subjective}. Language models give inconsistent answers to paraphrases of the same question \citep{elazar2021measuring}, their forecasts on logically related questions can be checked for consistency before outcomes are known \citep{paleka2025consistency}, and their probability judgments violate probability theory in ways that resemble human errors \citep{zhu2024incoherent}. Measurement in machine learning needs explicit validity checks \citep{jacobs2021measurement}. Verbalized confidences of language models can be better calibrated than their token probabilities \citep{tian2023just}, but elicitation struggles on questions that need expert knowledge \citep{xiong2024can}. Our score credits a true claim by its improbability under a panel's calibrated prior, not by the belief change of Bayesian surprise \citep{itti2009bayesian}, which AutoDiscovery uses as the reward for open-ended discovery \citep{agarwal2025autodiscovery}. Panels from several model families can replace a single judge \citep{verga2024replacing}; for surprise, we find that the panel's knowledge relative to the generator matters.

\paragraph{Reward overoptimization.}
Proxy--gold divergence is well documented for reward models and judges \citep{gao2022scaling,coste2023reward,zhou2026more,manheim2018categorizing,skalse2022defining,karwowski2023goodhart}. Optimizing for human approval can make models more convincing without making them more correct \citep{wen2025mislead}. Common-mode judge error is not identifiable from judge scores alone \citep{afrin2026evaluator}. Our gold signal is a measured scientific table.

\paragraph{Harness optimization and self-improvement.}
Optimizers search prompts, workflows and harnesses against a validation score \citep{agrawal2025gepa,hu2024automated,xu2026harness}, falling back on the model's own judgment when no verifier exists \citep{pan2026evolving}. Self-improving research agents rely on executable scores \citep{srikanth2026recursive,wang2026rsiforge} or on reviewer agents that check experimental evidence \citep{du2026rsimaster}. Self-refinement and debate do not correct errors without external feedback \citep{huang2023large,madaan2023self,du2023improving,laban2023sure}. Agents that run experiments optimize task performance or verify given hypotheses \citep{jiang2025aide,si2026towards,gandhi2025boxinggym,huang2025automated}. Our confirmed harness differs from these in that it uses no external feedback when it runs; its selection call is instructed to keep claims the generator believes but expects other models to doubt (whether it does so is examined in Appendix~\ref{app:selection}).

\section{Testbed}
\label{sec:setup}
\begin{figure}[t]
\centering
\begin{tikzpicture}[font=\small, box/.style={draw, rounded corners=2pt, align=center, inner sep=4pt, minimum height=1.0cm}, arr/.style={-{Stealth[length=2mm]}, thick}]
\node[box, text width=2.6cm] (h) {\textbf{Harness}\\retrieval, roles,\\challenge, selection,\\strategy text};
\node[box, text width=2.3cm, right=0.6cm of h] (p) {\textbf{Generator}\\(Sonnet / Opus / GPT)};
\node[box, text width=2.6cm, above=0.35cm of p] (m) {\textbf{Measurement}\\design $K$ probes,\\read results};
\node[box, text width=2.4cm, right=0.6cm of p] (y) {\textbf{10 hypotheses}\\``$j{=}b$ has a higher\\effect when $k$ is at $c$''};
\node[box, text width=2.5cm, right=0.6cm of y, yshift=0.95cm] (t) {\textbf{Table check}\\precision};
\node[box, text width=2.7cm, right=0.5cm of y, yshift=-1.05cm] (pr) {\textbf{Blind prior panel}\\leave-one-out\\$P$(higher, lower,\\no effect)};
\node[box, text width=1.9cm, right=0.5cm of t, yshift=-0.8cm] (sv) {\textbf{Surprise}\\$R$, \SV$_3$};
\node[box, text width=3.1cm, below=0.35cm of p] (o) {\textbf{Optimizer}\\fitness: LLM judge,\\\#true or \SV};
\draw[arr] (h) -- (p); \draw[arr] (m) -- (p); \draw[arr] (p) -- (y);
\draw[arr] (y) -- (t); \draw[arr] (y) -- (pr); \draw[arr] (t) -- (sv); \draw[arr] (pr) -- (sv);
\draw[arr] (o) -| (h); \draw[arr, dashed] (y) |- node[pos=0.25, right, font=\scriptsize] {scored traces} (o);
\end{tikzpicture}
\caption{Testbed. Every hypothesis is checked against a fully tabulated benchmark. Surprise is scored against a blind panel of other models that rated every possible claim: $R$ counts true hypotheses beyond what the calibrated panel expected, and \SV$_3$ counts true hypotheses the panel rated below 50. The panel was first asked only for $P(\text{higher})$ (original metric, Section~\ref{sec:s}) and from stage 12 on for all three outcomes.}
\label{fig:setting}
\end{figure}
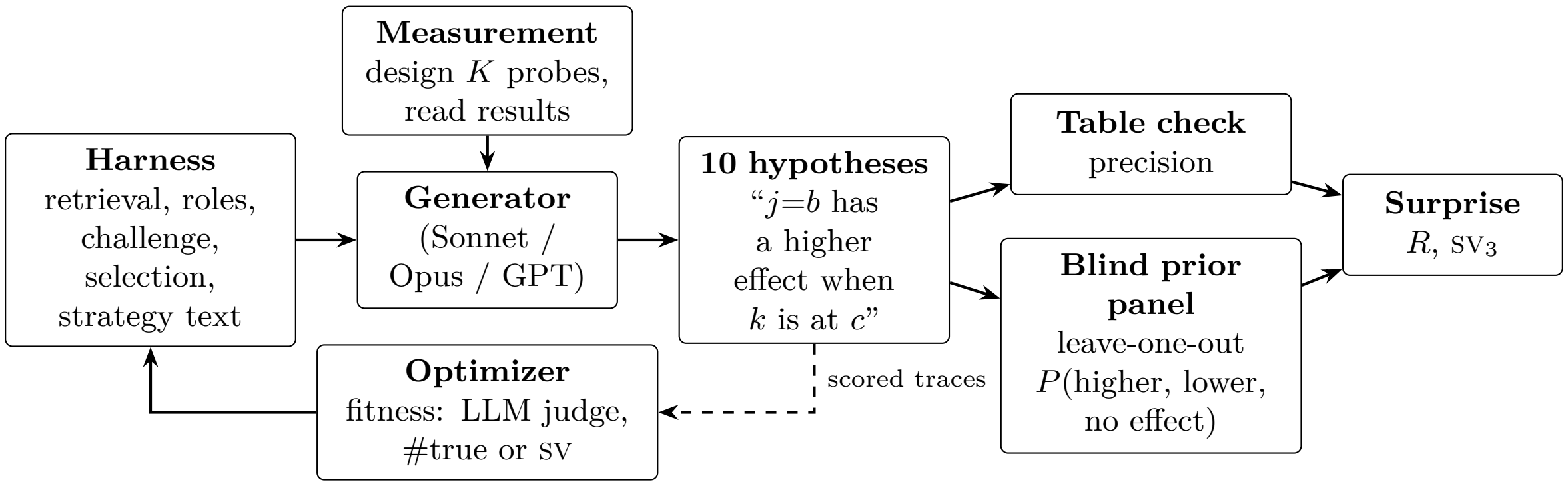

Figure~\ref{fig:setting} summarizes the testbed. Table~\ref{tab:stages} lists the stages, numbered in the order they were pre-registered, and the notation; Table~\ref{tab:ledger} gives every pre-registered criterion and its verdict.

\begin{table}[t]
\centering\footnotesize
\begin{tabular}{lp{12.6cm}}
\toprule
Stage & What it tested \\
\midrule
1 & Screen of structural harness knobs (NB201, Sonnet, 72 runs) \\
2, 3 & Harness optimization with three fitness functions (judge score, number of true claims, \SV); 3 reran the \SV{} optimizer after a bug fix \\
4, 4b & Measurement pilots (8 and 16 runs per arm) \\
5 & 24-run tests on NB201: judge-optimized harness and measurement (Sonnet, Opus) \\
6--8 & Measurement controls for Opus on NB201: simulated noise and a cheap proxy (6), shuffled or withheld values (7), designing without results (8) \\
9 & Replication on HPOBench (Sonnet, Opus) \\
10 & GPT as generator on both benchmarks \\
11 & Prompt follow-up on how Sonnet uses evidence \\
12, 12b & Three-way prior and corrected metric; GPT added to the panel \\
13 & Judge without explanations; 24-run confirmation of the stage-3 harness; token-matched direct baseline for GPT \\
14 & Selection-rule variants (on GPT's direct pools; ablation of the stage-3 harness); judge retest; GPT as second judge \\
15 & New direct runs for Sonnet; shuffled and withheld values for GPT; a single-call 40-claim direct baseline for GPT \\
\midrule
\multicolumn{2}{p{14.2cm}}{\emph{Arms.} Direct: one call for ten hypotheses. Measure-first: design 100 probes, read their results, then propose. Judge-optimized: the harness evolved for the judge's score (stage 2). Stage-3 harness: the harness from the rerun \SV{} optimizer. \emph{Scores.} Precision: share of a run's distinct claims that are true. $R$: true claims beyond what the calibrated panel expected, per run. \SV$_3$: true claims with corrected prior below 50; \SV: the same under the original prior. \emph{Models.} H/S/O/G: Haiku, Sonnet, Opus, GPT.} \\
\bottomrule
\end{tabular}
\caption{Stages and notation. NB201: NAS-Bench-201; HPO: HPOBench.}
\label{tab:stages}
\end{table}

\paragraph{Checkable hypothesis spaces.}
A claim names an edit (set parameter $j$ to value $b$), a condition (parameter $k$ at level $c$) and a direction. It says that the edit's effect on the metric is higher (or lower) when the condition holds than when it does not, on every dataset. The conditional difference is the edit's mean effect over the parents that satisfy the condition minus its mean effect over those that do not; Appendix~\ref{app:defs} gives the formula, the derivation of the counts below and a worked example. A run's precision is the share of its distinct, parseable claims that are true.
\begin{itemize}
\item \textbf{NAS-Bench-201} (NB201; \citealp{dong2020bench}). The searched cell's six edges and five operations give 870 (edit, condition) cells and 1,740 directional claims. A claim is true when the conditional difference has the same sign on CIFAR-10, CIFAR-100 and ImageNet16-120, with a minimum magnitude $\tau$ of 2 percentage points (pp). Of the claims, 198 are true (random precision .114).
\item \textbf{HPOBench} (HPO; \citealp{eggensperger2021hpobench}). The tabular MLP benchmark has five ordinal hyperparameters and 30,000 configurations, each trained with five seeds; we use three OpenML tasks. Edits move one grid step up or down, and conditions are low/mid/high bands. There are 148 cells and 296 claims, of which 41 are true with $\tau=0.5$ pp (random precision .139).
\end{itemize}
The thresholds and the all-datasets rule were fixed before any generation; Appendix~\ref{app:sens} varies both.

\paragraph{Blind prior panel.}
Haiku, Sonnet and Opus rate every cell blind, giving $P(\text{higher})$, $P(\text{lower})$ and $P(\text{no consistent, meaningful difference})$, summing to 100 (stage 12). A claim's prior is the panel's mean probability of its stated direction, over the panel members other than the generator. This leave-one-out panel has two members for a Claude generator and all three Claude models for GPT; stage 12b adds GPT as a fourth member. A panel's AUC is the area under the ROC curve of its prior for the stated direction against truth, over every claim in the space. On HPOBench, when $k\neq j$, ``moving $j$ up has a higher effect'' and ``moving $j$ down has a lower effect'' are the same proposition, so these wordings are merged before scoring.

\paragraph{Verified surprise.}
We score a run by $R=\sum_h (T_h-\pi_h)$ over its distinct claims (duplicates and, on HPOBench, equivalent wordings merged; Appendix~\ref{app:checks} also reports $R$ per distinct claim). Here $T_h$ is truth, and $\pi_h$ is the panel's prior calibrated to truth over the whole space, separately for each generator's panel. The calibration is a lightly ridge-regularized logistic model ($\lambda=10^{-3}$) on the prior's logit (prior clipped to $[0.5, 99.5]$), the claim's order (whether the condition is on the edited parameter itself, $k=j$), their interaction, and the claim's direction. The order and direction terms absorb most base-rate differences between claim types: self-conditions hold about three times as often as all other claims on NAS-Bench-201 (.275 against .088) and twice as often on HPOBench (.23 against .12), so a run gains little by choosing types alone. $R$ is counted in claims per run, and $R>0$ means the run's claims hold more often than the panel would have predicted. Because $R$ can rise when a run trades many likely claims for a few unlikely true ones, we read a gain in $R$ as an improvement only when precision stays at least 0.8 times the comparison arm's (Section~\ref{sec:disagree}). This bound is part of the pre-registered criteria of the measurement tests (stages 2, 4--6 and 8--10), the stage-12 rescoring and the stage-13 confirmation of the stage-3 harness. The same-cost and control comparisons of stages 13c, 14 and 15 were pre-registered on $p$ alone, so for them the bound is a post hoc reading rule; Appendix~\ref{app:checks} quotes the criteria. Post hoc, Appendix~\ref{app:decomp} splits $R$ into credit for true claims, $\sum_{\text{true}}(1-\pi_h)$, and the cost of false ones, $\sum_{\text{false}}\pi_h$, and gives every verdict at bounds from 0.6 to 1.0; the verdicts on the stage-3 and judge-optimized harnesses do not depend on the bound. We also report \SV$_3$, the number of true claims with prior below 50. $R$ was introduced post hoc in stage 12 and is pre-registered from stage 12b on; \SV$_3$ was pre-registered in stage 12. Both were computed on cached runs. Appendix~\ref{app:sens} varies the ridge penalty, clipping, covariates and fitting set.

\paragraph{What $R$ does and does not measure.}
$R$ is a measure of excess truth relative to a calibrated LLM panel, and we use ``surprise'' only in that sense. It is not surprise to experts or to the literature, which we did not measure. Four properties limit what a gain in $R$ shows. It depends on the panel, so the same claims can score differently against another panel (Section~\ref{sec:n}). It depends on the truth threshold: a claim that misses the threshold by a little costs its full $\pi$, so part of a gain can come from avoiding near misses rather than from finding unexpected truths, which is why we report the credit and cost terms separately at each threshold. It gives partial credit, $1-\pi$, even for true claims the panel expected, so a run of many likely true claims can score like a run of a few unlikely ones; the precision bound guards against the latter trade. And it inherits any misfit of the calibration, which Appendix~\ref{app:sens} varies. The credit term is the part closest to the intended construct, and we state which findings rest on it.

\paragraph{Harnesses, optimizer and measurement.}
A harness sets retrieval (none, in-domain, cross-domain), roles (single, or gap finder, innovator and critic), self-challenge, pool size and selection rule, and a free-text strategy of at most 150 words. Direct generation, the baseline, uses none of these. An Opus optimizer reads scored traces and proposes edits over 3 rounds of 3 candidates, each scored on 6 runs. Its fitness is the judge's score (Opus rating novelty and quality, each 1--10), the number of true hypotheses, or the original \SV{} (Section~\ref{sec:s}). In measure-first arms the generator first designs $K=100$ probes, each a parent configuration plus one edit, and reads the change in the metric. On NAS-Bench-201 this is a table lookup on the test split. Truth uses validation, but the two splits agree closely (edit effects correlate at $r\geq.997$, and 99.8\% of claims get the same truth label), so these probes are near-exact samples of the quantity truth averages (Appendix~\ref{app:checks}). On HPOBench it is the test-accuracy difference of a single random training seed. On the generators' own probes, its sign agrees with the 5-seed truth 70\% of the time when the true difference is at least 0.5 pp.

\paragraph{Budgets.}
Direct generation used 3.5--5.9k tokens per run, the measure-first arm 17.4--36.5k, the stage-3 harness 17.5k and the judge-optimized harness 24.2--35.1k (Table~\ref{tab:tokens} in the appendix). During optimization, candidate harnesses whose mean cost exceeded 40k tokens per run were disqualified. Claude counts include the CLI's own system prompt, about 4k tokens per call, and exclude failed retries. Probes are table lookups and cost no compute. Where the budget could matter we compare with arms of similar cost: for the stage-3 harness, an ablation with the same prompt and pool (16.2k); for GPT, pooled and 40-claim direct baselines (stages 13--15); for Opus, arms that see shuffled or no values. Every run proposes ten hypotheses, so precision and the surprise scores compare what a fixed-size output contains. After duplicates and, on HPOBench, equivalent wordings are merged, a run holds 9--10 distinct claims on NAS-Bench-201 and 6--10 on HPOBench.

\paragraph{Protocol.}
The study has 17 pre-registered stages, each with written decision criteria (16 documents; stage 4b is an addendum to stage 4). We use one-sided permutation tests on runs, Holm correction within a family, and 90\% bootstrap intervals. Key arms get 24 fresh runs. The models are claude-haiku-5-5, claude-sonnet-5-5 and claude-opus-5-5, called through the Claude Code CLI with tools disabled between 2026-10-21 and 2026-10-24 (UTC); the CLI's session logs report exactly these identifiers for every logged call on every date (Table~\ref{tab:versions}). The CLI logs also hold 1,504 earlier sessions, all of those on 2026-10-20 and 114 on 2026-10-21, that predate the first pre-registration (\texttt{PREREG\_s1.md}, 2026-10-21 00:56 UTC). They are pilots on other claim spaces and unrelated work, and are not used as evidence. Apart from two requests to review an unrelated manuscript that mentions NAS-Bench-201, a keyword search finds neither benchmark in them; the first session that poses a claim about a benchmark starts at 01:05 UTC, nine minutes after that document (Appendix~\ref{app:checks}). Stages 10, 12b, 13, 14 and 15 used gpt-6-astra, requested by explicit identifier, through codex-cli 0.162.0 (medium reasoning effort, read-only sandbox, empty working directory, 2026-10-21 to 23), as a generator, as a fourth panel member and, in stage 14, as a second judge. The study used over 4,500 successful model calls, all cached; the sensitivity analyses of Appendix~\ref{app:sens} use no new calls.

\section{Results}

\subsection{A selection step that raises verified surprise}
\label{sec:sel}
The harness accepted when the \SV{} optimizer was rerun in stage 3 proposes a pool of 30 claims. Its strategy asks for the generator's probability and a note on how an intuitive reader would judge each claim, but the output format has no field for a probability and none of the 709 pooled claims states one. A separate call then keeps the 10 that best satisfy a rule the optimizer wrote, including both thresholds: those the generator rates at least 55\% likely but expects other models to rate below 50\% (Appendix~\ref{app:prompts}; Appendix~\ref{app:defs} gives the two calls, what the optimizer saw, and the stage-2 bug fix that preceded this run). Under the original metric it had regressed on fresh runs (1.67 in training, 1.25 fresh), but a post hoc rescoring of all 19 harness evaluations singled it out as the one harness that raised $R$ (Section~\ref{sec:h}; $+1.98$, Holm $p=.02$, 8 runs). Three pre-registered tests followed, each with 24 fresh runs and the criterion $p<.05$; stage 13 also required precision at least 0.8 times direct generation's (Table~\ref{tab:sel}). The harness's precision was 1.05, 1.12 and 1.00 times the comparison arm's in the three tests. The three tests share the same 24 harness runs from stage 13 and differ only in the comparison arm, so they are one confirmation against three comparators, not three independent replications.

\begin{table}[t]
\centering\small
\setlength{\tabcolsep}{3pt}
\begin{tabular}{llccccccc}
\toprule
Comparison & Stage & Prec. & \SV$_3$ & $\Delta R$ [90\% CI] & $\Delta(R/n)$ & Calibrations & Panels & Truth rules \\
\midrule
Stage-3 harness & -- & .83 & 5.17 & & & & & \\
vs direct generation & 13 & .79 & 3.54 & $+1.77$ [$+1.30$, $+2.25$] & $+0.177$ & 14/14 & 8/8 & 6/6 \\
vs new direct runs & 15 & .74 & -- & $+1.95$ [$+1.45$, $+2.45$] & $+0.195$ & 14/14 & -- & 6/6 \\
vs likely-true ablation & 14 & .82 & 3.58 & $+1.96$ [$+1.55$, $+2.38$] & $+0.196$ & 14/14 & -- & 6/6 \\
\bottomrule
\end{tabular}
\caption{The stage-3 harness, Sonnet on NB201, 24 runs per arm, leave-one-out Claude panel (Haiku and Opus). All three tests pre-registered; $p<.001$ each. Prec.: precision of the comparison arm (first row: the harness). The ablation keeps the harness's strategy and pool of 30 but selects the 10 claims most likely to be true. Every arm holds 10 distinct claims per run (ablation 9.96); $\Delta(R/n)$: difference per distinct claim (post hoc; $p<.001$ each; Table~\ref{tab:perclaim}). Post hoc robustness, counting $p<.05$: Calibrations, 14 calibration variants (Table~\ref{tab:sens_cal}); Panels, the 8 distinct panels, namely the seven drawn from Haiku, Opus and GPT (the three-member one is the leave-one-out panel with GPT added) and Sonnet's own blind ratings (Table~\ref{tab:sens_panel}); Truth rules, six definitions of truth (Table~\ref{tab:sens_truth}). Table~\ref{tab:crossed} adds intervals that also resample propositions.}
\label{tab:sel}
\end{table}

\paragraph{Confirmation.}
Against direct generation (stage 13), $R$ rose by 1.77 per run, \SV$_3$ from 3.54 to 5.17 ($p<.001$), and precision from .79 to .83; the 8-run estimate ($+1.98$) shrank only to $+1.77$. Because that direct control had been run two days earlier, stage 15 ran 24 new direct runs: the harness beat them by 1.95, and the two direct batches differed by $-0.18$ ([$-0.64$, $+0.31$]). The harness used 17.5k tokens per run, against 5.2k for direct generation.

\paragraph{The selection rule carries the gain.}
A pre-registered ablation in stage 14 kept the strategy and the pool of 30 but selected the 10 claims most likely to be true. The gain vanished: $R$ fell 1.96 below the confirmed harness and $0.19$ below direct generation, at the same precision (.82), with \SV$_3$ at 3.58 and 16.2k tokens per run. The ablation also matched the new stage-15 direct batch ($-0.01$). Generating 30 claims with the same strategy at the same cost thus does not produce the gain; the selection rule does. This ablation favors the harness by construction, however: claims selected for being likely true have high priors and earn little credit under $R$, and the ablation did worse than random selection from its own pools. The random-selection contrast below is therefore the cleaner evidence that the selection step carries the gain.

\paragraph{What the rule selects.}
Post hoc, we compared the rule with other ways of keeping 10 claims from the same 24 cached pools of 30 (Appendix~\ref{app:selection}, 5,000 draws each). Random selection from the pools gives $R=+1.32$ per run, close to direct generation ($+1.10$); the rule's selections give $+2.87$, above every random draw. Keeping the ``Other'' type first (interactions between edges that both carry an operation other than \texttt{none}; Table~\ref{tab:mech}) gives $+1.92$, and the rule beats this type filter in every draw as well ($+0.95$). The rule doubles the Other type's share (20\% of kept hypotheses against 10\% of the pools and 3\% for direct generation), and these hold 77\% of the time against a base rate of 7\%; but it also raises precision within each claim type (self-conditions .93 kept against .78 not kept, Other .77 against .64). The ablation's likely-true selection did worse than random selection from its own pools ($-0.36$), mainly because it kept no Other-type claim, although its pools held 76, 35 of them true. With indicators for the edited operation and the condition's operation added to the calibration (the +Type column of Table~\ref{tab:sens_cal}; these are operation indicators, not the three claim types), the harness's gain over direct generation is $+1.73$. With the three claim types themselves in the calibration (an indicator for claims whose condition removes an edge, and its interaction with the prior), the three gains are $+1.73$, $+1.90$ and $+1.89$ (post hoc, all $p<.001$; Appendix~\ref{app:checks}). The judge-optimized harness moved to the same claim type (74\% of its hypotheses) but at 43\% overall precision (Section~\ref{sec:a}). The rule's design intent can also be imitated with cached priors (post hoc; Appendix~\ref{app:selection}). Keeping, from each pool, the claims the scoring panel rates least likely gives $+3.42$, and keeping first those Sonnet's blind prior rates at least 55 and the panel below 50 gives $+3.47$; both beat the harness ($+2.87$), but both read the very priors that $R$ subtracts, which the harness never sees. Keeping the claims with the largest gap between Sonnet's blind prior and the panel's gives $+1.88$, and keeping those Sonnet's blind prior rates most likely gives $+0.35$. The harness thus recovers about three quarters of the gain from random selection to a selector given the panel's priors, without access to them.

\paragraph{What the rule does not do.}
Two checks qualify the account in the rule's own words. First, compliance: measured with Sonnet's separately elicited blind prior, the kept claims are not the ones Sonnet rated at least 55\% likely (27\% of kept claims against 45\% of those not kept), and that blind prior does not discriminate truth within the pools (AUC .51). They are, however, enriched for claims the leave-one-out panel rated below 50 (67\% against 54\%). Only 5 of 24 selection replies mention other models at all. Whatever the generator does in context, its selections favor claims the panel doubts and are more often true, but we cannot show that they follow its own stated probabilities. Second, concentration: the harness's 240 hypotheses contain 93 distinct propositions. The largest single contribution comes from a false claim that direct generation proposes often and the panel expects (setting edge 2 to a $3{\times}3$ convolution helps less when input edge 0 is \texttt{none}; $\pi=.97$), which the harness avoids. That claim holds in the stated direction on all three datasets but misses $\tau$ on CIFAR-10 by 0.07 pp. Dropping the 10 propositions that contribute most from both arms removes the gain against the stage-5 direct runs ($+0.07$) and the ablation ($-0.02$), and leaves $+0.48$ ($p=.02$) against the new stage-15 direct runs; counting each distinct claim once per arm leaves $+0.60$ per run against direct generation. To put this concentration into the uncertainty, we resampled propositions jointly with runs (Table~\ref{tab:crossed}). The 90\% intervals widen from [$+1.30$, $+2.25$] to [$+0.38$, $+3.30$] against direct generation, and the gain stays above zero against all three arms under the main panel ($p_\times=.015$, $.004$, $.024$) and the panel with GPT added ($p_\times\leq.009$). Scored against Sonnet's own blind prior it does not ($p_\times=.116$, $.062$, $.192$). The gain is therefore real on fresh runs, and under the main panel it generalizes beyond the particular propositions sampled, but it rests on a few recurring ones, as one would expect from a 1,740-claim space that a generator samples repeatedly.

\paragraph{Surprising truths or avoided errors?}
Post hoc, we split the gain into credit for true claims and cost avoided on false ones (Appendix~\ref{app:decomp}; summarized in Section~\ref{sec:intro}). Against direct generation, $+1.04$ of the $+1.77$ is credit. Over runs, credit alone is significant against all three comparison arms ($+1.04$, $+1.13$ and $+1.22$; $+0.66$ to $+1.09$ under Sonnet's own ratings; all $p\leq.001$). With propositions resampled as well it is only marginal against direct generation ($p_\times=.054$; .028 and .045 against the other two arms) and not significant under Sonnet's own ratings (Table~\ref{tab:crossed}). The harness's true claims have a lower mean calibrated prior than direct generation's (.58 against .70). Most of the cost avoided comes from near misses, claims whose stated direction holds on every dataset but by less than $\tau$ (0.44 per run against 0.97 for direct generation), including the top contributor above. At $\tau=1$ and 1.5 pp, where most near misses become true, the gain is almost all credit and is significant with propositions resampled ($p_\times\leq.013$); at $\tau=3$ it splits evenly.

\paragraph{Robustness.}
Post hoc, the gain held under all seven panels drawn from the other three models ($+1.26$ to $+1.82$, including each single member and the three-member panel, which is the leave-one-out panel with GPT added, $+1.78$) and under Sonnet's own blind ratings ($+1.25$; all $p<.001$ over runs, but $p_\times=.12$ against Sonnet's ratings with propositions resampled). It held under all 14 calibration variants we tried, including a ridge penalty of 100 ($+1.01$) and a calibration fitted only on the two arms' own claims ($+0.79$), and on the uncalibrated excess-truth score ($+0.98$; Table~\ref{tab:sens_cal}). It held at every truth threshold from 1 to 3 pp and when truth requires agreement on two of three datasets or in the mean ($+1.08$ to $+1.84$ against the two direct batches, all $p<.001$; Table~\ref{tab:sens_truth}); at $\tau=3$ pp, where only 108 claims (6.2\%) are true, it is $+1.59$ and $+1.84$, and the ablation contrast $+1.91$, with precision above direct generation's (.63 against .59). The harness was found in one of two optimizer runs (the first returned direct generation) and tested only with Sonnet on NAS-Bench-201. Applied to GPT's pooled direct claims, the same rule raised $R$ over likely-true selection on NAS-Bench-201 ($+1.08$) but not on HPOBench ($-0.23$; pre-registered as descriptive in stage 14).

\subsection{GPT uses informative evidence, but no budget-efficient gain was shown}
\label{sec:meas}
For GPT, the measure-first arm designs 100 probes, reads their values and then proposes. On the same designed probes, it beat shuffled and withheld values on both benchmarks (stage 15, pre-registered; Table~\ref{tab:gpt}). The probes and the proposal call are identical across these arms; only what the values say differs, so GPT's proposals use what the probes say, not only the act of designing experiments or the longer prompt. This does not show how GPT combines the probes with its own knowledge, which our conflict analysis (Section~\ref{sec:p}) examined only for Sonnet and Opus.

\begin{table}[t]
\centering\small
\setlength{\tabcolsep}{3pt}
\begin{tabular}{lllcccccc}
\toprule
Bench. & Comparison (stage) & Tokens & $n$ & $\Delta R$ [90\% CI] & $\Delta(R/n)$ & Prec.\ ratio & Verdict & Calibrations \\
\midrule
NB201 & Pooled direct (13) & 21.2k & 10.0 & $+2.24$ [$+1.79$, $+2.74$] & $+0.224$ & .71 & Supported & 12/14 \\
 & Pooled + stage-3 rule (14) & -- & 10.0 & $+1.16$ [$+0.57$, $+1.76$] & $+0.116$ & .68 & Supported & 12/14 \\
 & 40-claim call (15) & 11k & 9.8 & $+0.88$ [$+0.38$, $+1.40$] & $+0.086$ & .61 & Supported & 12/14 \\
 & Shuffled values (15) & -- & 10.0 & $+1.40$ [$+0.91$, $+1.92$] & $+0.140$ & 2.64 & Supported & 14/14 \\
 & Withheld values (15) & -- & 10.0 & $+0.89$ [$+0.39$, $+1.41$] & $+0.089$ & .72 & Supported & 11/14 \\
HPO & Pooled direct (13) & 20.4k & 7.5 & $+0.70$ [$+0.39$, $+1.02$] & $+0.081$ & .98 & Supported & 12/14 \\
 & Pooled + stage-3 rule (14) & -- & 5.5 & $+0.94$ [$+0.53$, $+1.35$] & $+0.055$ ($p=.08$) & 1.15 & Supported & 12/14 \\
 & 40-claim call (15) & 11.0k & 7.3 & $+0.09$ [$-0.31$, $+0.49$] & $-0.005$ ($p=.44$) & .97 & Refuted & 0/14 \\
 & Shuffled values (15) & -- & 8.5 & $+1.42$ [$+0.96$, $+1.89$] & $+0.181$ & 1.63 & Supported & 14/14 \\
 & Withheld values (15) & -- & 9.0 & $+0.85$ [$+0.45$, $+1.26$] & $+0.127$ & 1.05 & Supported & 12/14 \\
\bottomrule
\end{tabular}
\caption{GPT, measure-first minus each pre-registered comparison arm on $R$, 24 runs per arm, leave-one-out Claude panel. Measure-first costs 21.0k tokens per run on NB201 and 17.4k on HPO. Supported: Holm $p\leq.010$ within the stage; refuted: Holm $p=.36$ with the 90\% CI upper bound below 0.5. Calibrations: number of the 14 calibration variants of Table~\ref{tab:sens_cal} under which the difference has $p<.05$ (post hoc). Measure-first precision: .50 on NB201 (direct .69) and .95 on HPO (direct .90). Prec.\ ratio: measure-first precision over the comparison arm's; the 0.8 bound was not part of these stages' pre-registered criteria, so the verdicts follow $p$ alone, but under our reading rule the NAS-Bench-201 gains over every same-cost arm (ratio below 0.8) do not count as improvements. $n$: distinct claims per run in the comparison arm (measure-first: 10.0 on NB201, 7.8 on HPO). $\Delta(R/n)$: difference per distinct claim (post hoc; $p\leq.005$ unless shown; Table~\ref{tab:perclaim}).}
\label{tab:gpt}
\end{table}

\paragraph{Against direct generation and same-cost baselines.}
Stage 10 compared measurement with direct generation, pre-registered on the original metric: on HPOBench it raised \SV{} by 1.08 (Holm $p<.001$) without losing precision (.89 to .94); on NAS-Bench-201 the gain ($+0.79$, Holm $p=.003$) failed the precision condition (.69 to .50). Post hoc, $R$ rises by 0.94 and 1.60 per run. Because measurement costs about five times the tokens of direct generation, stages 13--15 added baselines of similar cost (Appendix~\ref{app:checks}). Measurement beat four pooled direct runs, with or without the stage-3 rule, on both benchmarks, and a single 40-claim call on NAS-Bench-201; on HPOBench the 40-claim call matched it at about two thirds of its tokens (11.0k against 17.4k). Per distinct claim (post hoc; Table~\ref{tab:perclaim}), one more HPOBench gain weakens: against pooled direct generation with the stage-3 rule, which keeps only 5.5 distinct claims per run, it is $+0.055$ ($p=.078$).

\paragraph{Scope.}
On HPOBench the gain over direct generation holds under all 14 calibrations, the uncalibrated score and every Claude panel, and it keeps precision, but a cheaper call matches it. On NAS-Bench-201 it costs precision beyond the bound and is calibration-dependent: with a ridge penalty of 100, a calibration fitted on the two arms' own claims, or the uncalibrated score, measurement no longer beats direct generation ($-0.46$, $-0.67$, $-0.45$), although it still beats shuffled values. There, moreover, a probe reads almost exactly the quantity that truth averages (Section~\ref{sec:setup}), so measurement is close to sampling the answer. On neither benchmark have we shown that measurement is a budget-efficient way to raise $R$.

\subsection{A measurement gain for Opus depends on the panel (post hoc, hypothesis-generating)}
\label{sec:n}
\paragraph{A non-Claude panel member.}
Stage 12b asked GPT the same three-way question (34 calls). GPT puts less mass on ``no effect'' (Table~\ref{tab:bias}), and its prior predicts truth at least as well as Opus's (AUC .91 and .95; Opus .88 and .93, Sonnet .71 and .72, Haiku .67 and .69; 90\% intervals within $\pm.04$ on NAS-Bench-201 and $\pm.09$ on HPOBench). With GPT added to the leave-one-out panel (pre-registered), measurement raised both corrected scores for Opus on NAS-Bench-201 (\SV$_3$ $+1.42$, $R$ $+0.90$, Holm $p<.05$ over the four Claude pairs), with precision .87 to .75, within the bound. This gain does not come from the original artifact (Section~\ref{sec:s}). GPT rates most of Opus's directly generated true claims as likely, mainly claims that setting another edge to \texttt{none} changes an edit's effect (93 of the 109 true-claim instances over 24 runs that leave \SV$_3$), and measurement moves Opus away from such claims (from about two thirds to .47 of hypotheses). Under a GPT-only panel, $R$ favors measurement for Opus on NAS-Bench-201 by $+1.25$ (Holm $p=.002$). On the other Claude pairs the verdicts were inconclusive (Sonnet on NAS-Bench-201, Opus on HPOBench) or refuted (Sonnet on HPOBench; Table~\ref{tab:m}).

\paragraph{Panel dependence.}
Post hoc, we rescored measurement against direct generation under every panel drawn from the models other than the generator (Figure~\ref{fig:panel}; details in Appendix~\ref{app:panel}). On NAS-Bench-201, measurement raises $R$ for Opus and GPT under all 8 panels that include the other frontier model (AUC $\geq .87$; $+0.84$ to $+1.71$, $p\leq.015$, uncorrected) and under none of the 6 that do not (AUC $\leq .74$; $-0.56$ to $+0.30$). Sonnet gains under no panel, including its own blind ratings. On HPOBench, GPT gains under all seven panels and Opus under four. Against Opus's own blind ratings, a strong panel without the other frontier model, Opus gains too ($+0.91$ on NAS-Bench-201, $+0.48$ on HPOBench), and the same probes with shuffled or withheld values do not produce its gain (Table~\ref{tab:controls}). We read this as a hypothesis, not a finding. No panel has an AUC between .74 and .87, so panel quality and the presence of the other frontier model are confounded. The boundary was read off the data, and the panels share members, so the split is one contrast per generator. Under the main Claude panel, Opus's gain is $+0.19$ and significant in none of the 14 calibrations, and with GPT in the panel it disappears under heavy shrinkage ($-0.03$ with a ridge penalty of 100). A pre-registered test with a panel member of intermediate AUC would separate the explanations.

\begin{figure}[t]
\centering
\includegraphics[width=0.62\linewidth]{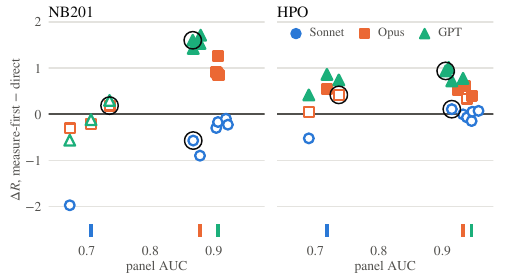}
\caption{Measure-first minus direct generation on $R$ against the panel's AUC, for every panel drawn from the other three models (post hoc, 24 runs per arm). Filled: $p<.05$, one-sided, uncorrected. Circled: the Claude leave-one-out panel of Table~\ref{tab:m}. Ticks along the bottom: each generator's own AUC.}
\label{fig:panel}
\end{figure}

\subsection{A precision check shows the cost of optimizing for judged novelty}
\label{sec:a}
We evolved one harness for the judge's score, in one optimizer run, so this section is a case study of that harness. Table~\ref{tab:a} shows that it raises that score and cuts precision by 46--68\%, also when transferred unchanged to HPOBench and when GPT is the generator (pre-registered in stages 5, 9 and 10). The evolved strategy forbids two claim types ``because experts find them obvious'': conditioning on the current value of the edited parameter, and conditions that make the edited component unused. These are the claim types most often true (Table~\ref{tab:mech}). Before any optimization, a screen of 72 runs had shown what the judge rewards: its novelty rating correlates with the mean prior of a run's hypotheses at $\rho=-.58$ ($-.74$ under the corrected prior, post hoc), and its quality rating with the number of true hypotheses at $\rho=+.53$. Post hoc, the precision loss holds under every truth definition of Appendix~\ref{app:sens}: for all three generators and all eleven definitions, the judge-optimized harness falls below 0.8 times direct generation's precision (Table~\ref{tab:sens_truth}).

\paragraph{Type shift or worse claims?}
The judge prompt asks whether experts would find the hypotheses surprising, and the evolved strategy bans the two claim types most often true, so part of the precision loss follows from base rates. Post hoc, we compared precision within claim type (Table~\ref{tab:judgetype}). On HPOBench every judge-optimized claim is of the Other type, and these hold far less often than direct generation's Other-type claims (.35--.43 against .80--.96; the 90\% intervals of the difference all lie below $-0.3$). The type mix alone would predict a precision of .80--.96, so on HPOBench the harness's claims are worse within type, not only of a different type. On NAS-Bench-201 direct generation proposes almost no Other-type claims (8, 6 and 0 over 24 runs), so the two explanations cannot be separated there; within the one type both arms use often, the judge-optimized harness is less precise for Sonnet and Opus (.49 against .65, .52 against .89) and similar for GPT (.62 against .66). All of this concerns one judge-optimized harness from one optimizer run; we have not shown that judge optimization in general behaves this way.

\paragraph{Style or content?}
Three pre-registered rejudgings of all 288 direct and judge-optimized runs ask whether the judge's gain is a matter of style (Table~\ref{tab:judge}; details in Appendix~\ref{app:more}). Without each hypothesis's one-sentence ``why'', the judge still preferred the judge-optimized sets in four of six pairs, always through novelty ($+0.62$ to $+1.42$), not quality. A retest with the original prompt reproduced the gain ($+0.46$ to $+1.92$), but single-run totals barely correlate between judgings ($r=-.22$ to $.62$), so only arm means reproduce. A second judge, GPT, rated the judge-optimized sets higher in novelty and lower in quality in all six pairs, so its total rose only slightly (0 to $+0.33$; pre-registered verdict partial). Both judges thus read the harness's claims as more novel; precision, which needs no judge, shows what they cost.

\paragraph{$R$ alone would not have caught it.}
Under the corrected prior, the judge harness lowers $R$ by 0.9--2.2 per run in three of four Claude pairs and for GPT on HPOBench ($p\leq.001$; with a prior-only calibration, significantly only in the two Opus pairs). For Sonnet and GPT on NAS-Bench-201, however, it raises $R$ ($+0.88$, $+1.04$) while precision falls to .43 and .26. Half or more of the rise is cost avoided, because its many false claims are ones the panel also doubted (credit $+0.43$ and $-0.38$, cost avoided $+0.45$ and $+1.41$; Appendix~\ref{app:decomp}); for GPT this rise turns into a drop under a weaker Haiku and Sonnet panel ($-0.86$). This is why we pair $R$ with the precision bound (Section~\ref{sec:disagree}).

\begin{table}[t]
\centering\small
\setlength{\tabcolsep}{3pt}
\begin{tabular}{lllcccc}
\toprule
Bench. & Gen. & Harness & Prec. & Judge & \SV$_3$ & $\Delta R$ \\
\midrule
NB201 & Sonnet & Direct & .79 & -- & 3.54 & \\
 & & Judge-opt. & .43 & -- & 4.25 & $+0.88^{\dagger}$ \\
 & Opus & Direct & .87 & 8.25 & 6.96 & \\
 & & Judge-opt. & .28 & 8.88 & 2.75 & $-2.24^{\dagger}$ \\
 & GPT & Direct & .69 & 8.04 & 4.08 & \\
 & & Judge-opt. & .26 & 8.63 & 2.58 & $+1.04^{\dagger}$ \\
HPO & Sonnet & Direct & .82 & 8.12 & 5.33 & \\
 & & Judge-opt. & .43 & 9.17 & 4.33 & $-0.89^{\dagger}$ \\
 & Opus & Direct & .90 & 8.17 & 5.38 & \\
 & & Judge-opt. & .42 & 9.96 & 4.25 & $-1.69^{\dagger}$ \\
 & GPT & Direct & .89 & 8.33 & 3.33 & \\
 & & Judge-opt. & .35 & 8.63 & 3.50 & $-0.87^{\dagger}$ \\
\bottomrule
\end{tabular}
\caption{Judge fitness, 24 fresh runs per arm (stage 5 for NB201, stage 9 for HPO, stage 10 for GPT). The harness was evolved with Sonnet on NB201 (stage 2) and applied unchanged to the other generators and to HPO. Judge: novelty + quality (2--20); Sonnet's NB201 runs in stage 5 were not judged (stage 13 judged them without explanations). Pre-registered: precision drop $\geq .2$ with $p<.05$ in every judge-optimized row; judge increase $p<.05$ in every row with judge scores; both held. HPO precision is unmerged; merging equivalent wordings changes it by at most .01. $\Delta R$: change in claims true beyond the calibrated leave-one-out Claude panel's expectation (post hoc); $^{\dagger}p<.01$, one-sided in the direction shown, uncorrected.}
\label{tab:a}
\end{table}

\begin{table}[t]
\centering\small
\setlength{\tabcolsep}{3pt}
\begin{tabular}{lccc}
\toprule
 & Self-cond. & Removes edge & Other \\
\midrule
Base rate of truth (all claims) & .275 & .173 & .067 \\
Share, Sonnet direct / judge-opt. & .68 / .03 & .28 / .23 & .03 / .74 \\
Share, Opus direct / judge-opt. & .31 / .00 & .67 / .17 & .03 / .82 \\
\bottomrule
\end{tabular}
\caption{Claim types on NB201, as shares of all hypotheses of 24 runs per arm (about 240; a claim proposed in several runs counts each time). The judge-optimized strategy forbids self-conditions and conditions that make the edited component unused. ``Removes edge'' counts any condition that sets another edge to \texttt{none}, which is broader than the ban, so such claims remain at .17--.23. The stage-3 harness puts 20\% of its hypotheses in ``Other'', and 77\% of those are true.}
\label{tab:mech}
\end{table}

\begin{table}[t]
\centering\small
\setlength{\tabcolsep}{3pt}
\begin{tabular}{llccccccc}
\toprule
 & & \multicolumn{4}{c}{Opus judge, $\Delta$ total} & \multicolumn{3}{c}{GPT judge} \\
\cmidrule(lr){3-6}\cmidrule(lr){7-9}
Bench. & Gen. & Original & No ``why'' (13) & Retest (14) & ``Why'' effect & $\Delta$ total & $\Delta$ novelty & $\Delta$ quality \\
\midrule
NB201 & Sonnet & -- & $+1.17$ & $+0.67$ & $-0.50$ & $+0.04$ & $+0.83$ & $-0.79$ \\
 & Opus & $+0.62$ & $+0.42$ & $+0.62$ & $+0.21$ & $0.00$ & $+1.00$ & $-1.00$ \\
 & GPT & $+0.58$ & $+1.46$ & $+0.67$ & $-0.79^{\dagger}$ & $+0.33^{*}$ & $+1.08$ & $-0.75$ \\
HPO & Sonnet & $+1.04$ & $+0.21$ & $+1.42$ & $+1.21^{\dagger}$ & $+0.29^{*}$ & $+0.88$ & $-0.58$ \\
 & Opus & $+1.79$ & $+0.71$ & $+1.92$ & $+1.21^{\dagger}$ & $+0.33^{*}$ & $+0.88$ & $-0.54$ \\
 & GPT & $+0.29$ & $+0.38$ & $+0.46$ & $+0.08$ & $+0.33$ & $+0.88$ & $-0.54$ \\
\bottomrule
\end{tabular}
\caption{Judge-optimized minus direct generation, judge totals (novelty + quality, 2--20; components 1--10), 24 runs per arm. Original: the stage 5, 9 and 10 judgments. No ``why'': explanations removed (stage 13; kept in four pairs, $p\leq.022$). Retest: the original prompt, judged again (stage 14). ``Why'' effect: retest gain minus no-``why'' gain, two-sided, Holm over six pairs; $^{\dagger}$Holm $p<.05$. GPT judge: the original prompt with GPT as judge (stage 14); $^{*}$one-sided $p<.05$ (pre-registered ``kept''). Every GPT novelty and quality difference has $p\leq.001$ (post hoc).}
\label{tab:judge}
\end{table}

\subsection{Sensitivity of the positive results}
\label{sec:robust}
Appendix~\ref{app:sens} reports three post hoc sensitivity analyses on cached runs, added after the main study. \emph{Calibration} (Tables~\ref{tab:sens_cal} and~\ref{tab:sens_cal_own}): 14 variants of the ridge penalty ($0$ to $100$), prior clipping ($0.1$ to $5$), covariates (prior only; claim-type indicators), 10-fold cross-fitting by cell, and a fit on the two compared arms' claims only, plus the uncalibrated $\sum(T-p/100)$. Clipping changes no difference by more than 0.02 and cross-fitting by no more than 0.09. The stage-3 harness and GPT's and Opus's advantages over shuffled values are significant in all 14 variants under both the leave-one-out panel and the generator's own ratings; GPT's other NAS-Bench-201 gains and Opus's other gains are not robust to $\lambda=100$ or the arms-only fit (details in Appendix~\ref{app:sens}). \emph{Truth definition} (Table~\ref{tab:sens_truth}): $\tau\in\{1,1.5,2,3\}$ pp on NAS-Bench-201 and $\{0.25,0.5,1\}$ pp on HPOBench, and agreement on two of three datasets or in the mean instead of on all three. The stage-3 harness's gain and GPT's measurement gain over direct generation hold under every definition (at $\tau=3$ on NAS-Bench-201, $+1.59$ and $+1.34$; at $\tau=1$ on HPOBench, GPT $+0.75$), although GPT's NAS-Bench-201 precision stays below the bound throughout. At these two strictest thresholds the default Newton fit of the calibration diverges, so we refit them with step halving, which reproduces the default fit wherever that converges (all 1,657 fits of the calibration and panel analyses). The judge harness fails the precision bound everywhere. \emph{Panel size} (Table~\ref{tab:sens_panel}): see Section~\ref{sec:n}. The stage-3 harness is the result that survives every analysis in the appendix, although once propositions are resampled it does not survive scoring against Sonnet's own prior (Table~\ref{tab:crossed}); the measurement results for GPT on NAS-Bench-201 and for Opus depend on the calibration as well as the panel.

\section{Diagnostics: what did not work and why}
\label{sec:diag}

\subsection{A complement-coded surprise metric manufactured a gain}
\label{sec:s}
Before stage 12, each panel member gave only the probability that the edit's effect is \emph{higher} under the condition, and a lower claim's prior was 100 minus it. A hypothesis was surprising-valid (\SV) if true with prior below 50. If the panel's beliefs over the three outcomes are $P(\text{higher})$, $P(\text{lower})$ and $P(\text{none})$, the complement rule gives a lower claim the prior
\[ 100-P(\text{higher}) = P(\text{lower})+P(\text{none}) \geq P(\text{none}). \]
When the panel expects no effect for most cells, a lower claim can count as surprising only if the panel confidently expected the opposite direction. The same holds whenever one direction's probability is derived from a one-sided answer, for example by taking 100 minus the rated plausibility of ``X improves Y'' as that of ``X harms Y''.

\begin{figure}[t]
\centering
\includegraphics[width=\linewidth]{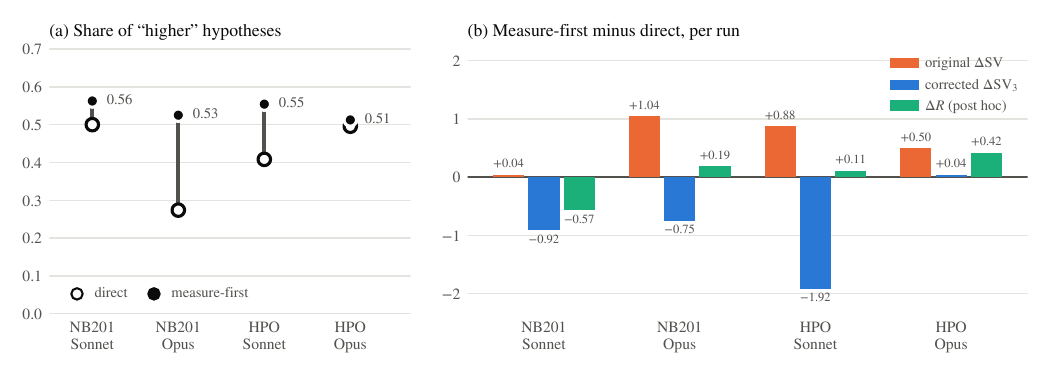}
\caption{The direction bias, Claude generators and Claude panel. (a) Share of higher-direction hypotheses, the only direction the original metric could credit; Opus on HPO barely changes. (b) Measure-first minus direct generation under the original metric, the corrected metric (\SV$_3$, pre-registered rescoring) and the threshold-free calibrated score $R$ (post hoc), 24 runs per arm.}
\label{fig:dir}
\end{figure}

Under this metric, stages 5, 6 and 9 confirmed that measurement raises \SV: Opus on NAS-Bench-201 from 1.08 to 2.12 per run (Holm $p=.003$), Sonnet on HPOBench from 1.83 to 2.71 and Opus from 4.17 to 4.67. An external audit then pointed to the elicitation; its numbers reproduced exactly from our caches (Table~\ref{tab:bias}, Figure~\ref{fig:dir}). The panel's mean $P(\text{higher})$ was 21--46, so no true lower claim was \SV-eligible in any of the four model--benchmark pairs, while 57\% (NAS-Bench-201) and 54\% (HPOBench) of true claims are lower-direction. On HPOBench the two wordings of a proposition received opposite labels for 41--60 of 60 propositions. Measurement shifted proposals toward higher claims, the only direction the metric credited (Opus on NAS-Bench-201 from .27 to .53 of hypotheses, Sonnet on HPOBench from .41 to .55), and this shift accounts for $+1.02$ of Opus's $+1.04$ and $+0.68$ of Sonnet's $+0.88$ (Table~\ref{tab:m}).

Stage 12 re-elicited a three-way prior (102 calls). Its pre-registration was written before any stage-12 call but after the audit, and the runs it rescored had already been analyzed, so it constrains the analysis but not the data. All three models put the largest share on ``no effect'' (47--70 on average); the two HPOBench wordings now differ by a median of 3 points; and 92\% and 95\% of true lower claims have a prior below 50. Under \SV$_3$ the measurement effect for Claude generators is refuted in three pairs and inconclusive in the fourth, and the pre-registered surprisal-weighted credit agrees. Keeping the original ratings for higher claims and using the new ones only for lower claims (pre-registered) gives the same picture ($-0.42$, $-0.88$, $-1.12$, $-0.04$), so the reversal comes from the missing lower-direction prior, not from the new question format. The stage-3 harness's gain was measured on the corrected metrics and does not depend on this artifact.

\begin{table}[t]
\centering\small
\setlength{\tabcolsep}{3pt}
\begin{tabular}{lcc}
\toprule
 & NB201 & HPO \\
\midrule
Mean $P(\text{higher})$, H/S/O/G & 21/25/46/54 & 22/28/40/49 \\
Lower share of true claims & .57 & .54 \\
Lower share of \SV-eligible true claims & .00 & .00 \\
Equivalent wordings, opposite labels & -- & 41--60/60 \\
Three-way $P(\text{none})$, H/S/O/G & 70/68/47/24 & 68/69/53/38 \\
True lower claims, prior $<50$ (three-way) & .92 & .95 \\
\bottomrule
\end{tabular}
\caption{The direction bias of the complement-coded surprise metric. H/S/O/G: Haiku/Sonnet/Opus/GPT; probabilities are means over cells. GPT's two-way ratings (stage 10) are descriptive and enter no panel. \SV-eligible: true and leave-one-out prior $<50$ (averaged over both generators' panels).}
\label{tab:bias}
\end{table}

\begin{table}[t]
\centering\small
\setlength{\tabcolsep}{3pt}
\begin{tabular}{llcccccc}
\toprule
 & & & & \multicolumn{2}{c}{Corrected, Claude panel} & \multicolumn{2}{c}{GPT added to panel} \\
\cmidrule(lr){5-6}\cmidrule(lr){7-8}
Bench. & Gen. & Orig.\ $\Delta$\SV{} & Mix part & $\Delta$\SV$_3$ [90\% CI] & $\Delta R$ & $\Delta$\SV$_3$ & $\Delta R$ \\
\midrule
NB201 & Sonnet & $+0.04$ & $+0.14$ & $-0.92$ [$-1.42$, $-0.42$] & $-0.57$ & $+0.29$ & $-0.10$ \\
 & Opus & $+1.04^{*}$ & $+1.02$ & $-0.75$ [$-1.42$, $-0.04$] & $+0.19$ & $+1.42^{\ddagger}$ & $+0.90^{\ddagger}$ \\
 & GPT & $+0.79$ & $+0.11$ & $+0.08$ [$-0.58$, $+0.75$]$^{\S}$ & $+1.60^{\dagger}$ & -- & -- \\
HPO & Sonnet & $+0.88^{*}$ & $+0.68$ & $-1.92$ [$-2.33$, $-1.50$] & $+0.11$ & $-1.33$ & $+0.05$ \\
 & Opus & $+0.50^{*}$ & $+0.15$ & $+0.04$ [$-0.50$, $+0.62$] & $+0.42$ & $+0.08$ & $+0.52$ \\
 & GPT & $+1.08^{*}$ & $+0.25$ & $+0.67$ [$+0.25$, $+1.12$]$^{\S}$ & $+0.94^{\dagger}$ & -- & -- \\
\bottomrule
\end{tabular}
\caption{Measure-first minus direct generation, 24 fresh runs per arm. Precision as pre-registered: NB201 .79$\to$.45 (Sonnet), .87$\to$.75 (Opus), .69$\to$.50 (GPT); HPO .82$\to$.83, .90$\to$.89, .89$\to$.94 (with merged wordings, as used for \SV$_3$ and $R$: .81$\to$.82, .90$\to$.89, .90$\to$.95). $^{*}$Pre-registered support under the original metric; GPT on NB201 was inconclusive because precision fell below 0.8 times direct generation's. Mix part: the part of the original difference explained by the change in the number of higher-direction claims, holding \SV{} per higher claim at the two arms' mean (post hoc). Corrected, Claude panel: the leave-one-out Claude panel (two members for Claude generators, three for GPT). Under \SV$_3$ (pre-registered rescoring, stage 12) the four Claude pairs are refuted, refuted, refuted, inconclusive; $^{\S}$post hoc. For Claude generators no Claude-panel $\Delta R$ is significant in the positive direction (smallest $p=.06$); $^{\dagger}p<.001$, one-sided, post hoc. GPT added to panel: the Claude leave-one-out panel plus GPT (stage 12b), pre-registered; $^{\ddagger}$Holm $p<.05$ over the four Claude pairs. A GPT generator's leave-one-out panel cannot include GPT.}
\label{tab:m}
\end{table}

\subsection{Why the two corrected metrics disagree}
\label{sec:disagree}
For Claude generators with the Claude panel, $\Delta$\SV$_3$ is negative in three pairs while $\Delta R$ is positive, though not significant, in three. With three outcomes most claims get a prior below 50, so \SV$_3$ is close to a count of true, non-obvious claims and tends to fall when precision falls. $R$ credits each true claim by how unlikely the calibrated panel found it, so a shift toward lower-prior claims can offset a precision loss, and $R$ alone can reward a loss of truth: for GPT on NAS-Bench-201 it rewarded the judge-optimized harness, at 26\% precision, by $+1.04$ under the Claude panel and $+1.91$ under GPT's own prior, not far below measurement ($+1.60$, $+2.69$). Hence the pre-registered precision bound. Measurement meets it for Opus on both benchmarks and for GPT on HPOBench, and misses it for GPT and Sonnet on NAS-Bench-201 (.69 to .50, .79 to .45). The stage-3 harness meets it with room to spare (.79 to .83).

\subsection{Sonnet does not discount unrepresentative evidence}
\label{sec:p}
Sonnet gained from measurement under no panel. For Sonnet on NAS-Bench-201, measurement lowers precision from .79 to .45. Post hoc, we split measure-first hypotheses by whether the model's own probes point in the direction that its own blind three-way rating considered less likely of the two (Table~\ref{tab:conflict}). On NAS-Bench-201 such conflict claims rarely concern real effects (25--30\% of the time), because single-parent contrasts often reflect one parent architecture's idiosyncrasy. Of Sonnet's claims, 37\% are conflict claims, against 5\% of Opus's. A pre-registered prompt follow-up (stage 11; Appendix~\ref{app:more}) suggests that Sonnet discounts evidence described as noisy but not evidence described as unrepresentative: removing the ``single training seed'' wording on HPOBench raised its siding with conflicting evidence, while a truthful note on NAS-Bench-201 that each result comes from a single parent cell did not lower it. Asking Sonnet to propose only claims that its own knowledge also supports raised precision from .44 to .60 (Holm $p=.001$), still below direct generation. Two of the stage's four predictions missed their thresholds.

\begin{table}[t]
\centering\small
\setlength{\tabcolsep}{3pt}
\begin{tabular}{llccc}
\toprule
Bench. & Gen. & Conflict claims (share) & True-cell share & Precision: conflict / agree \\
\midrule
NB201 & Sonnet & 86 (.37) & .30 & .24 / .56 \\
 & Opus & 12 (.05) & .25 & .17 / .78 \\
HPO & Sonnet & 21 (.09) & .57 & .48 / .88 \\
 & Opus & 17 (.07) & .76 & .76 / .89 \\
\bottomrule
\end{tabular}
\caption{Measure-first hypotheses split by whether the model's own probes agree with the direction its own blind three-way rating prefers (post hoc, 24 runs per arm; hypotheses without probe evidence omitted). True-cell share: the share of conflict claims whose cell has a real effect in either direction. With the original one-directional rating the conflict shares are similar (Sonnet 42\% on NB201, Opus 15\%), but siding rates are not comparable across ratings (Sonnet on HPO: .61 against .10), because the two define the preferred direction differently.}
\label{tab:conflict}
\end{table}

\subsection{Other harnesses did not beat direct generation}
\label{sec:h}
We rescored, post hoc, all 19 harness evaluations on fresh NAS-Bench-201 runs, 13 for Sonnet and 6 for Opus (the judge harness was evaluated twice per model). They include an L9 orthogonal-array screen of the structural knobs, the final harnesses of the three fitness arms, prior-avoiding selection, cross-domain retrieval, and a harness that designs 100 experiments it knows will not be run. None beats direct generation on \SV$_3$ after Holm correction (Table~\ref{tab:harness} in the appendix). On $R$, two Sonnet harnesses do: the judge-optimized harness ($+0.88$, Holm $p=.04$, at precision .43) and the stage-3 harness ($+1.98$, Holm $p=.02$), which Section~\ref{sec:sel} confirms. No Opus harness significantly exceeds direct generation on either score. The validity fitness, also grounded in truth, produced a harness that played safe (Table~\ref{tab:ledger}). With 8 runs per arm for 15 of the 19 evaluations, the smallest \SV$_3$ difference detectable with 80\% power was about two claims per run (about three after Holm correction), so these are absence-of-evidence results. Cross-domain retrieval, the largest single gain reported for a recent ideation system \citep{liu2026ideascientist}, did not help on any metric.

\begin{table}[t]
\centering\footnotesize
\begin{tabular}{lp{6.8cm}p{7.4cm}}
\toprule
Stage & Pre-registered prediction & Verdict \\
\midrule
1 & Structural knobs (retrieval, roles, challenge, selection) shift \SV{} by $\geq 1$/run & Inconclusive \\
1 & Judge novelty does not track \SV{} ($\rho\leq.2$) & Supported ($\rho=-.24$; novelty vs prior $\rho=-.58$) \\
2 & Judge fitness raises the judge score without raising \SV & Supported ($p=.018$; precision .78$\to$.29) \\
2 & Validity fitness plays safe (prior $\uparrow\geq 3$, fewer distinct claims) & Supported (prior $+6.2$) \\
2 & \SV{} fitness beats the other two & Refuted (final harness = direct generation) \\
3 & No harness beats direct generation on \SV & Supported (the \SV{} rerun after a bug fix regressed on fresh runs); post hoc, \SV$_3$ agrees, but $R$ favors that rerun harness (8 runs; confirmed in stage 13) \\
3 & Prior and precision trade off across harnesses & Inconclusive \\
4, 4b & Measurement raises \SV; more probes raise it more & Inconclusive (8 and 16 runs/arm); the 8 new Opus runs alone: refuted ($+0.12$) \\
5 & Judge harness: precision $\downarrow\geq .2$, judge $\uparrow$, \SV{} $\downarrow$ & Precision and judge supported; \SV{} refuted / inconclusive \\
5 & Measure-first raises \SV{} (24 runs/arm) & Supported (Opus), refuted (Sonnet); \textbf{reversed in stage 12} \\
6 & Gain survives simulated single-seed noise / doubled noise / 12-epoch proxy & Supported / inconclusive / inconclusive \\
7 & Shuffled and withheld values: is the gain driven by the values or by the design? & Both inconclusive \\
8 & A design-only harness beats direct & Inconclusive (Opus), refuted (Sonnet) \\
9 & HPO: judge harness precision $\downarrow$, judge $\uparrow$ & Supported (both models) \\
9 & HPO: judge harness \SV{} $\downarrow$ & Supported (Opus), refuted (Sonnet) \\
9 & HPO: measure-first raises \SV & Supported (both); stage 12: \textbf{refuted} (Sonnet), inconclusive (Opus) \\
10 & GPT, judge harness: precision $\downarrow\geq .2$, judge $\uparrow$, \SV{} $\downarrow$ & Precision and judge supported, \SV{} refuted (both benchmarks) \\
10 & GPT, measure-first raises \SV & Inconclusive on NB201 (precision .69$\to$.50); supported on HPO ($+1.08$) \\
11 & Weak-evidence note raises NB201 precision by $\geq .15$ (Sonnet) & Inconclusive ($+.04$) \\
11 & Knowledge filter raises NB201 precision by $\geq .15$ & Supported ($+.15$, Holm $p=.001$) \\
11 & Removing the seed wording lowers HPO precision by $\geq .10$ & Inconclusive ($-.06$, Holm $p=.007$) \\
12 & $\geq 20\%$ of true lower claims are surprising under a three-way prior & Supported (92\%, 95\%) \\
12 & Measure-first raises \SV$_3$ & Refuted in three pairs, inconclusive for Opus on HPO \\
12b & Stage 12 holds with GPT in the panel & Not robust: Opus on NB201 gains ($+1.42$ \SV$_3$, $+0.90$ $R$) \\
13 & The judge gain survives removing each hypothesis's ``why'' (all six pairs) & Partial: kept in four ($+0.42$ to $+1.46$), weakened in two (HPO, Sonnet and GPT) \\
13 & The stage-3 \SV{} rerun harness raises $R$ (Sonnet, NB201, 24 runs), precision $\geq 0.8\times$ direct & Supported ($+1.77$, $p<.001$; precision .79$\to$.83) \\
13 & GPT: measure-first beats a token-matched direct baseline on $R$ & Supported on both benchmarks: $+2.24$ (NB201), $+0.70$ (HPO), Holm $p<.001$; pooling alone: $-0.64$, $+0.24$ \\
14 & GPT: measure-first also beats the baseline with the stage-3 harness's selection rule & Supported on both benchmarks: $+1.16$ (NB201), $+0.94$ (HPO), Holm $p\leq.002$ \\
14 & The stage-3 harness's gain comes from its selection rule (likely-true ablation, Sonnet, NB201) & Supported ($+1.96$, $p<.001$; ablation vs direct $-0.19$) \\
14 & Explanations carry part of the judge gain (retest with them, per pair) & In two pairs (HPO, Sonnet and Opus: $+1.21$, Holm $p<.001$); none on NB201, where GPT went the other way ($-0.79$) \\
14 & The judge gain transfers to a second judge, GPT (all six pairs) & Partial: kept in three ($+0.29$ to $+0.33$), weakened in two, vanished in one (NB201, Opus); novelty up and quality down in all six \\
15 & The stage-3 harness beats a current batch of direct runs (Sonnet, NB201) & Supported ($+1.95$, $p<.001$; drift between direct batches $-0.18$) \\
15 & GPT: measurement beats shuffled and withheld values (each benchmark) & Supported on both: $+1.40$, $+0.89$ (NB201), $+1.42$, $+0.85$ (HPO), Holm $p\leq.005$ \\
15 & GPT: measurement beats a single-call 40-claim direct baseline (each benchmark) & NB201 supported ($+0.88$, Holm $p=.010$); HPO refuted ($+0.09$, $p=.36$)
\\
\bottomrule
\end{tabular}
\caption{Every pre-registered decision criterion and its verdict. Each pre-registration was written before any call of its stage; stage 12's was written after the external audit and rescored runs already analyzed under the original metric. Differences are computed before rounding.}
\label{tab:ledger}
\end{table}

\section{Discussion}
\label{sec:disc}

\paragraph{Why might the selection rule work?}
The confirmed harness's rule, written by the optimizer, asks the generator in a separate selection call to keep the claims it believes but expects other models to doubt. Direct generation and likely-true selection keep the claims the generator is most confident about, which are also the claims every model expects. The intent is to spend the generator's own knowledge where it differs from the consensus, which is what $R$ rewards. In our data the rule does more than filter claim type: from the same pools it beat random selection by 1.55 and an Other-type filter by 0.95 per run, and it raised precision within each type. But the kept claims are not those Sonnet's separately elicited blind prior rates likely, so the mechanism in the rule's wording, belief against expected doubt, remains a hypothesis; what we can show is that the selection step favors claims the panel doubts and that hold. The rule needs no external feedback at run time, so it is cheap to apply. Its 50\% threshold echoes the score definition the optimizer was shown. We have tested it with only one generator on one benchmark, and on GPT's pooled direct claims it helped on one benchmark and not the other.

\paragraph{When does evidence help?}
For GPT, measurement raised $R$ on both benchmarks only when the probe values were informative, so its proposals use the evidence. We have not shown that measurement is a good use of budget: on HPOBench a cheaper 40-claim call matched it, and on NAS-Bench-201 it cost precision. Nor have we tested how GPT weighs evidence against its own knowledge. Measurement never helped Sonnet, which followed unrepresentative probes and lost precision. For Opus a gain appeared only under panels at least as informed as Opus. One hypothesis, for a pre-registered test with a panel member of intermediate quality, is that a gain from evidence can register as surprise only relative to a panel that knows at least as much as the generator.

\paragraph{Surprise relative to whom, and contamination.}
$R$ measures surprise relative to the blind priors of other LLMs, not relative to experts or the literature. Both benchmarks are public \citep{dong2020bench,eggensperger2021hpobench}, and the strongest priors predict truth with AUC .88 to .95, so the models may have seen the tables or papers about them. Contamination that the panel shares is discounted by construction, because the calibrated prior already predicts those claims. A generator that recalls more than its panel, however, earns $R$ for recall rather than reasoning, which is one reading of the panel dependence in Section~\ref{sec:n}. For the stage-3 harness this reading fits less well: Sonnet's own blind prior predicts truth less well than Opus's, with an AUC of .71 against .88; the gain is positive when scored against Sonnet's own prior, at $+1.25$ per run, although not above zero with propositions resampled; and the kept claims are not those that prior rates likely. We cannot rule contamination out. A private or post-cutoff table would, but the one private-table transfer we ran (Appendix~\ref{app:more}) tested only the judge-optimized harness.

\paragraph{Checks for evaluating ideation systems.}
Each failure we met has a cheap check that would have caught it.
\begin{itemize}
\item \emph{Report precision on a checkable subset next to any judge or surprise score.} Judge-scored benchmarks would rank the judge-optimized harness higher, and $R$ alone rewarded it in two pairs.
\item \emph{Elicit priors over every outcome, not one direction}, and audit the direction balance of the true claims a surprise metric can credit against that of all true claims. Ours was 0\% against 54--57\%, a table that needs no calls beyond the priors themselves.
\item \emph{Make the panel at least as informed as the generator.} Report each member's and the generator's own AUC against truth. A leave-one-out panel from one model family is weakest for that family's strongest model, which then earns credit for what it already knows. The generator's own blind ratings are a cheap strong panel, although they capture its knowledge only in part.
\item \emph{Merge logically equivalent claims before scoring}, and check that a proposition's label does not depend on its wording.
\item \emph{Report a calibrated excess-truth score with its sensitivity to the calibration.} Our GPT and Opus measurement gains on NAS-Bench-201 depended on how much the calibration trusted the prior; the stage-3 harness's did not.
\end{itemize}

\paragraph{What pre-registration did and did not protect.}
Pre-registration \citep{nosek2018preregistration} kept us from tuning analyses after seeing results, and it made the confirmation of the stage-3 harness and the reversal of the original measurement result equally easy to report. It did not protect the construct: the flawed metric was locked in before any data were collected and passed three confirmation stages. Construct checks such as a direction-balance audit need no calls beyond the priors and belong in the pre-registration, as a gate before any confirmatory stage.

\paragraph{Relation to recursive self-improvement.}
Harness evolution succeeds when selection uses executable truth \citep{srikanth2026recursive,wang2026rsiforge}, and reviewer agents keep experimental claims honest when scores exist \citep{du2026rsimaster}. Without an executable score, evolution here failed with a judge. With a sparse fitness that counted only true, low-prior claims, the original \SV{} that we later found to be biased by direction, one of two optimizer runs produced a harness that we singled out post hoc among 19 evaluations. For one generator on one benchmark, its gain held on corrected scores, fresh seeds and every panel we tried. It rests partly on a few recurring claims, however, and with those resampled it is not distinguishable from zero against the generator's own prior. The harness encodes the objective in its selection step, and replacing only that step removed the gain. Whether harnesses found this way transfer to other generators and benchmarks is the most important open question this work leaves.

\section{Limitations}
\begin{itemize}
\item The stage-3 harness was found in one of two optimizer runs and confirmed with one generator (Sonnet) on one benchmark (NAS-Bench-201). Its gain rests on a few recurring propositions: without the 10 that contribute most it is not significant against two of three comparison arms. The harness and its ablation were run a day apart; their pools are similar, with pool $R$ differing by $+0.19$ per run at $p=.76$, so we attribute little of their difference to the batch. With propositions resampled along with runs, its gain is above zero under the main panel but not under Sonnet's own blind prior, and its credit term is only marginal against direct generation ($p_\times=.054$).
\item Conditional-interaction claims on tabular benchmarks are a narrow proxy for research ideas. Both benchmarks are public, and the strongest models' priors predict truth with AUC .88--.95, so they may partly know the tables (Section~\ref{sec:disc}).
\item $R$ before stage 12b and all sensitivity analyses are post hoc, and $R$ depends on a calibration fitted to truth over the claim space. The measurement gains for GPT on NAS-Bench-201 and for Opus do not survive heavy shrinkage of that calibration, and per distinct claim GPT's HPOBench gain over pooled direct generation with the stage-3 rule is not significant. On NAS-Bench-201 the probes' test-split values almost equal the validation values that define truth. The three-way prior may itself be pulled toward an even split over its three outcomes.
\item The judge is Opus except in the stage-14 transfer test with GPT, and the prior panel is Claude except in stage 12b and the post hoc panel analyses. GPT ran through codex-cli, an agent shell whose behavior may differ from the API, and neither interface controls sampling temperature. Opus is the judge, the optimizer's proposer and a panel member, so self-preference may favor Opus outputs \citep{panickssery2024llm}; the leave-one-out panel excludes the generator but not its model family.
\item Stage 12 rescored runs that had already been analyzed, so its pre-registration constrains the analysis but not the data.
\item Stages 13 to 15 compare new arms with direct and measure-first runs made up to three days earlier. For Sonnet, 24 direct runs made alongside stage 15 differed from the stage-5 batch by $-0.18$ per run, and the stage-3 harness's gain held against them. For GPT, direct runs repeated one to two days later changed $R$ by at most 0.3 per run. Opus's direct runs varied more across batches (2.98 to 3.71), but no comparison in stages 13 to 15 uses an Opus generator.
\item In one of 24 Sonnet measure-first runs on NAS-Bench-201, all 100 probes failed to parse because the model abbreviated operation names. Excluding it leaves \SV{} at 1.17 and precision at .44.
\item The optimizer budget is small, and most harness comparisons have 8 runs per arm, so negative results for harnesses may partly reflect search limits and low power. The judge result rests on one judge-optimized harness from one optimizer run, evolved with Sonnet and transferred unchanged; it shows that judge optimization can cost precision, not how often it does.
\end{itemize}
The 16 pre-registration documents are listed with their SHA-256 hashes in Table~\ref{tab:prereg} and are included in the supplement and at \url{https://osf.io/c2hkg/?view_only=d05d8b1acd504e6ab87a242522b769df}; every model call on the two public benchmarks, the analysis scripts and a command-line tool for checkable hypothesis generation will be released.

\subsubsection*{Broader Impact Statement}
This work evaluates automated research-ideation systems. A reader could treat a gain on a narrow checkable subset as evidence that a system produces true and valuable ideas in general; our scores certify neither outside a checkable setting. Systems tuned to an LLM judge can produce plausible but false claims, as the judge-optimized harness did. An objective that rewards surprise relative to a panel rewards claims that contradict consensus, which is useful only when truth is checked; the selection rule we confirm should be paired with such checks. The study depends on closed models whose behavior may change. The public benchmarks contain no personal data.

\bibliography{refs}
\bibliographystyle{tmlr}

\clearpage
\appendix
\input{appendix}

\end{document}

%% file: appendix.tex
\section{Prompts}
\label{app:prompts}
All prompts are shown for NAS-Bench-201. The HPOBench versions use their own benchmark description, claim wording and JSON fields (edited parameter and step direction, conditioning parameter and band), and describe each probe as two configurations trained once with a single training seed. Text in angle brackets is filled in per call. Claude models were called through the Claude Code CLI with tools disabled, and GPT through codex-cli in a read-only sandbox in an empty working directory. Prompts for the harness roles, the stage-1 screen, the stage-11 variants and the optimizer are in the released code.

\paragraph{Benchmark description (prefix of every NAS-Bench-201 prompt).}\mbox{}\par\nobreak
\begin{lstlisting}
Benchmark: NAS-Bench-201 cell search space. A cell is a DAG with nodes 0 (input) to 3 (output) and six edges; each edge carries one operation.
Edges: edge 0 = node 0 -> node 1; edge 1 = node 0 -> node 2; edge 2 = node 1 -> node 2; edge 3 = node 0 -> node 3; edge 4 = node 1 -> node 3; edge 5 = node 2 -> node 3.
Operations: none, skip_connect, nor_conv_1x1, nor_conv_3x3, avg_pool_3x3 ('none' removes the edge).
Datasets: CIFAR-10, CIFAR-100, ImageNet16-120; fitness is validation accuracy of the full network built by stacking the cell.
An edit sets one edge j to operation b. A condition is 'parent edge k currently has operation c' (k may equal j, in which case it says what op b replaces).
A hypothesis states that the edit's effect on validation accuracy is HIGHER (or LOWER) when the parent satisfies the condition than when it does not, consistently across the three datasets and by a practically meaningful margin. It is about the interaction, not the edit's average effect.
\end{lstlisting}

\paragraph{Direct generation.}\mbox{}\par\nobreak
\begin{lstlisting}
<benchmark description>
(Independent attempt #<n>.)

Propose 10 conditional hypotheses about this benchmark that you believe are true.
End with JSON only: {"hypotheses": [{"edit_edge": <0-5>, "edit_op": "<op>", "cond_edge": <0-5>, "cond_op": "<op>", "direction": "higher" or "lower", "why": "<one sentence>"}, ...]} with exactly 10 distinct hypotheses.
\end{lstlisting}

\paragraph{Measure-first: design call, then proposal call.}\mbox{}\par\nobreak
\begin{lstlisting}
<benchmark description>
(Independent attempt #<n>.)

You may run 100 experiments before proposing hypotheses. An experiment applies one edit to one parent cell you choose and returns the change in test accuracy (percentage points) on the three datasets. Design the experiments that will best reveal conditional effects; you will see the results and then propose 10 hypotheses.
End with JSON only: {"experiments": [{"parent": ["<op edge 0>", ..., "<op edge 5>"], "edit_edge": <0-5>, "edit_op": "<op>"}, ...]} with exactly 100 experiments.
\end{lstlisting}
\begin{lstlisting}
<benchmark description>
(Independent attempt #<n>.)

Before proposing, you ran 100 real experiments on this benchmark. Each experiment applied one edit to one parent cell and measured the change in test accuracy (percentage points) on the three datasets; these are real measurements and can serve as evidence:
1. parent (e0=<op>, ..., e5=<op>); edit: edge <j> -> <op>; change C10 <+x.xx>, C100 <+x.xx>, IN16 <+x.xx>
...

Propose 10 conditional hypotheses about this benchmark that you believe are true.
<same JSON format as direct generation>
\end{lstlisting}

\paragraph{Original prior elicitation (one direction, complement-coded).} Each call rates 30 cells.\mbox{}\par\nobreak
\begin{lstlisting}
<benchmark description>

For each numbered claim, give your probability (0-100) that it holds (consistent across the three datasets, practically meaningful margin). If you think the effect is instead LOWER under the condition, give a low number.
0. setting edge <j> to <b> has a HIGHER effect when parent edge <k> is <c> than when it is not
...
End with JSON only: {"p": {"0": <0-100>, ...}} covering every claim.
\end{lstlisting}

\paragraph{Corrected prior elicitation (three outcomes, stage 12).} Each call rates 30 cells.\mbox{}\par\nobreak
\begin{lstlisting}
<benchmark description>

For each numbered (edit, condition) pair, compare the edit's effect on validation accuracy when the condition holds with its effect when it does not. Give three probabilities (0-100, summing to 100): H = the effect is HIGHER under the condition, L = the effect is LOWER under the condition, N = neither (no difference that is consistent across the three datasets and of practically meaningful size).
0. edit: set edge <j> to <b>; condition: parent edge <k> is <c>
...
End with JSON only: {"p": {"0": [H, L, N], ...}} covering every pair.
\end{lstlisting}

\paragraph{Judge.} The judge sees each hypothesis as JSON, including its one-sentence ``why''; stage 13a repeats the judge with that field removed, and stage 14 repeats it unchanged with Opus and with GPT as the judge.\mbox{}\par\nobreak
\begin{lstlisting}
<benchmark description>

A research assistant proposed these hypotheses:
{"hypotheses": [{"edit_edge": 4, "edit_op": "nor_conv_3x3", "cond_edge": 4, "cond_op": "none", "direction": "higher", "why": "<the generator's one-sentence reason>"}, ...]}

Rate this set as a reviewer would, without running anything: novelty (1-10: would an expert find these surprising and non-obvious?) and quality (1-10: soundness, specificity, plausibility). End with JSON only: {"score": {"novelty": <1-10>, "quality": <1-10>}}
\end{lstlisting}

\paragraph{The evolved judge-optimized harness.} Roles: gap finder, innovator and critic; challenge: negate assumptions; pool of 20, reduced to 10 by the rule below; no retrieval.\mbox{}\par\nobreak
\begin{lstlisting}
Selection rule: Discard any hypothesis whose condition is the edited field's own current value, or whose condition merely disables or disconnects the edited component. From the rest, keep 10 with distinct mechanisms and field pairs, favouring those where an informed expert could plausibly bet on either direction.

Strategy: Two patterns are banned because experts find them obvious: (1) conditioning on the current value of the same field being edited, which just restates that upgrading a weak setting helps more than upgrading a strong one; (2) conditions that make the edited component unused, unreachable or inert. Instead, each hypothesis must link two DIFFERENT fields that interact through a non-trivial mechanism, such as substitution between parallel routes, a capacity already supplied elsewhere, one part acting as a bottleneck for another, or interference that turns a usually helpful edit harmful. No condition setting may appear in more than two hypotheses. Use the gaps and negated assumptions as the source of each idea. In each 'why', name the mechanism and explain why the opposite direction is also a priori plausible.
\end{lstlisting}

\paragraph{The harness evolved for verified surprise (stage 3, confirmed in stage 13).} Roles: single; no challenge; pool of 30, reduced to 10 by the rule below; no retrieval.\mbox{}\par\nobreak
\begin{lstlisting}
Selection rule: Keep hypotheses you rate at least 55% likely true but expect other models to rate below 50%, and name the specific naive reasoning that would mislead them. Drop anything you rate below 50% yourself, and anything obvious from structure alone. Keep fields and directions diverse.

Strategy: Propose a broad, varied set of conditional hypotheses covering many different edited fields, conditioning fields and both directions. For each, briefly note your own probability that it holds and how an intuitive reader would judge it.
\end{lstlisting}

\section{Additional results}
\label{app:more}
\paragraph{Judge harness on a private table.}
A descriptive transfer to a private, unreleased table of model results from another domain, pre-registered in stage 2, showed the same pattern as Table~\ref{tab:a} (judge 6.63 to 8.75, precision .40 to .03; 8 runs per arm). About 3\% of that table's directional claims are true, so the judge-optimized harness fell to chance precision there. We report it for completeness and do not rely on it; its runs are not part of the release. For GPT as generator, the original judge gains were $+0.58$ on NAS-Bench-201 and $+0.29$ on HPOBench ($p=.005$ and $.04$).

\paragraph{Judge rejudgings in detail.}
The judge sees each hypothesis with its one-sentence ``why'', and the evolved strategy prescribes how to write it (``name the mechanism and explain why the opposite direction is also a priori plausible''). Stage 13 removed the ``why'' and left the prompt otherwise unchanged. The judge still preferred the judge-optimized sets in four of six pairs (verdict partial), and in every pair the gain was in novelty ($+0.62$ to $+1.42$), not in quality. Stage 14 repeated the original prompt. The retest reproduced the gain ($+0.46$ to $+1.92$). Only for Sonnet and Opus on HPOBench did the explanations add to it, by 1.21 points each (Holm $p<.001$). For Sonnet and Opus on NAS-Bench-201 we found no significant effect, and for GPT on NAS-Bench-201 removing the explanations raised the gain by 0.79 (Holm $p=.012$), which we did not predict. Within an arm, a run's original and retest totals barely correlate ($r=-.22$ to $.62$), so single-run judge scores are unreliable and only arm means reproduce. Stage 14 also gave the original prompt to GPT, a transfer test, since the harness was evolved against Opus. GPT rated the judge-optimized sets higher in novelty and lower in quality in all six pairs (post hoc, $p\leq.001$ for each), so its total rose only slightly (0 to $+0.33$). The rise was significant in three pairs, so the pre-registered verdict is partial. GPT's ratings were nearly binary: novelty was 2 for 134 of 144 direct sets and 3 for 137 of 144 judge-optimized sets. When GPT judged GPT's runs, both arms were its own, so self-preference would affect both.

\paragraph{Measurement variants and threshold sweep.}
With the Claude panel, every 24-run measurement variant for Opus on NAS-Bench-201 scores below direct generation on \SV$_3$ (6.96): exact values 6.21, simulated single-seed noise 5.92, doubled noise 5.88, values withheld 5.88, a 12-epoch proxy 5.25, and values shuffled across probes 4.58. A threshold sweep counts only true claims the panel rated below 33 (pre-registered as descriptive) or 20 (post hoc). For Opus, measurement gains $+0.92$ at 20 and $+0.79$ at 33 on HPOBench, and $+0.12$ and $+1.00$ on NAS-Bench-201 ($p\leq.01$ except $+0.12$, $p=.40$). With GPT in the panel (post hoc), the gains at 33 are $+0.88$ and $+0.75$ ($p<.01$), but at 20 they are $+0.00$ and $+0.04$, because that panel rates almost no true claim below 20.

\paragraph{How the models use conflicting evidence.}
When the models make conflict claims (Table~\ref{tab:conflict}), they follow the probes on NAS-Bench-201 (92\% for Sonnet, 83\% for Opus, pooled over claims) but less often on HPOBench (10\%, 24\%), where probes are described as single training runs. A pre-registered follow-up (stage 11) changed only the propose prompt and reused Sonnet's own probes (23--24 paired runs per arm); its conflict measures used the original one-directional rating. Removing the ``single training seed'' wording on HPOBench raised Sonnet's siding with conflicting evidence from .61 to .80 and lowered precision by .06 (Holm $p=.007$; short of the pre-registered .10). A truthful note on NAS-Bench-201 that each result comes from a single parent cell did not lower siding (.91 to .95; predicted below .75) and changed precision by only $+.04$. Asking Sonnet to propose only claims that its own knowledge also supports raised precision from .44 to .60 (Holm $p=.001$), still below direct generation, but cut conflict claims only from 40\% to 30\% (predicted below 25\%). Recomputed post hoc with the three-way rating of Table~\ref{tab:conflict}, which defines the preferred direction differently, the shifts are .11 to .42, .91 to .89, and 37\% to 26\%; under neither rating does the note lower siding toward the predicted .75. The contrast is confounded with the benchmark and with adding versus removing text, and neither the two prompt-wording precision predictions nor the two mechanism predictions reached their thresholds.

\input{appendix_selection}

\input{appendix_sensitivity}

\input{appendix_definitions}

\input{appendix_decomp}

\input{appendix_panel}

\input{appendix_checks}

\section{Additional tables}
\label{app:tables}
Tables~\ref{tab:tokens}--\ref{tab:self} are generated from the cached analysis outputs by a script released with the paper.

\input{appendix_tables}

%% file: appendix_selection.tex
\section{Selection analyses of the stage-3 harness}
\label{app:selection}
All analyses in this section are post hoc and use only the cached generation pools, selections, runs and priors of the stage-3 harness (stage 13) and its likely-true ablation (stage 14), Sonnet on NB201, 24 runs each (script \texttt{post\_pos\_selection.py}). $R$ uses the calibration of the main text (leave-one-out Claude panel). A pool holds 29.5 distinct valid claims on average, and every kept claim is in its pool.

\paragraph{Selection baselines.}
Each baseline keeps as many claims from each pool as the arm kept (10), and is repeated 5,000 times (Table~\ref{tab:selbase}). ``Other first'' keeps claims of the Other type (Table~\ref{tab:mech}) before any other, in random order, and fills up at random; ``non-self first'' does the same for every claim that does not condition on the edited edge. The draw distribution holds the pools fixed, so it measures the selection step, not the sampling of pools; the paired test compares each run's kept $R$ with that run's expected baseline $R$ (500 draws) and flips signs over runs.

\begin{table}[h]
\centering\small
\setlength{\tabcolsep}{3pt}
\begin{tabular}{llcccc}
\toprule
Pools & Selection & $R$/run [5\%, 95\% of draws] & Prec. & Kept $-$ baseline & Paired: runs kept $>$, $p$ \\
\midrule
Stage-3 harness & Rule (kept) & $+2.87$ & .825 & & \\
 & Random & $+1.32$ [$+0.92$, $+1.72$] & .745 & $+1.55$ & 24/24, $<.001$ \\
 & Other first & $+1.92$ [$+1.57$, $+2.27$] & .739 & $+0.95$ & 22/24, $<.001$ \\
 & Non-self first & $+1.45$ [$+1.18$, $+1.73$] & .646 & $+1.42$ & \\
 & Other, then removes-edge & $+1.62$ [$+1.35$, $+1.89$] & .648 & $+1.25$ & \\
Ablation & Likely-true (kept) & $+0.91$ & .824 & & \\
 & Random & $+1.26$ [$+0.89$, $+1.66$] & .749 & $-0.36$ & 7/24, .96 \\
 & Other first & $+1.55$ [$+1.20$, $+1.88$] & .683 & $-0.64$ & 8/24, .99 \\
\bottomrule
\end{tabular}
\caption{Selection from the same pools of 30. No baseline draw reached the rule's $R$. Direct generation (stage 5) has $R=+1.10$ per run.}
\label{tab:selbase}
\end{table}

\paragraph{Proxies for ``believed but doubted''.}
The rule asks for claims the generator believes and expects others to doubt. Post hoc, we imitated that intent with the cached blind priors, keeping from each pool as many claims as the harness kept, in a fixed order with random tie-breaking (200 tie-breaks; script \texttt{post\_pos\_round6b.py}). Keeping the claims the leave-one-out panel rates least likely gives $R=+3.42$ per run (precision .66). Keeping first the claims that meet the rule's thresholds under blind priors (Sonnet's own $\geq 55$, the panel's $<50$; 3.5 per pool, 12\% of pool claims), then the rest by panel prior, gives $+3.47$ (.70); keeping the lowest calibrated priors gives $+3.38$ (.65). Each beats the harness's own selections ($+2.87$, precision .83; paired sign-flip $p\geq.989$ for the harness being better). These selectors, however, read the panel's priors, which are exactly what $R$ subtracts, and the harness never sees them. Selectors that use Sonnet's blind prior do worse: the largest gap between Sonnet's blind prior and the panel's gives $+1.88$ (harness better by $+0.99$, $p=.003$), and Sonnet's most confident claims give $+0.35$ ($+2.52$, $p<.001$). Measured against random selection ($+1.32$) and the panel-informed ceiling ($+3.42$), the harness recovers about three quarters of what is available, at higher precision than any panel-informed selector.
Finally, we scored each pooled claim from generator-side signals only, without reading the scoring panel's priors: in-context selection support combined with the generator's own blind probability $P_{\text{own}}(c)$, minus a ridge-fitted structural prior $\hat{P}_{\text{struct}}(c)$ (script \texttt{post\_pos\_excess\_truth.py}). This selector is post hoc and its weights were set by hand after the results above were known; it also selects from pools of 30 or more claims, whereas a direct run proposes 10. It raised $R$ over direct generation in all six benchmark--generator pairs, but significantly in only two: Sonnet on NAS-Bench-201 ($+0.68$, $p=.008$) and GPT on HPOBench ($+1.18\to+2.06$, $p=.0003$, precision $.898\to.917$), the pair where the stage-3 prompt rule gave no gain over its comparison arm. The other four differences are $+0.14$ to $+0.43$ ($p=.07$--$.20$). We read this as a hypothesis for why the prompt rule did not transfer, not as evidence.

\paragraph{Claim types.}
Shares of claim instances (precision): harness pools self-condition .59 (.83), removes-edge .31 (.59), Other .10 (.72); harness kept .57 (.93), .23 (.62), .20 (.77); ablation pools .62 (.82), .27 (.70), .11 (.46); ablation kept .68 (.89), .32 (.68), .00; direct generation, stage 5, .68 (.87), .28 (.65), .03 (.25); stage 15, .63 (.87), .29 (.59), .08 (.21). Within each type, the claims the rule kept were more often true than those it left (self .93 against .78, removes-edge .62 against .59, Other .77 against .64) and had higher mean $T-\pi$ ($+0.22$ against $+0.07$, $+0.32$ against $-0.01$, $+0.46$ against $+0.35$).

\paragraph{Compliance.}
We measure the rule's two thresholds with priors elicited separately from the runs: Sonnet's own blind three-way prior ($\geq 55$) and the leave-one-out Claude panel's prior ($<50$). Of the kept claims, 27\% have an own prior of at least 55 (pool claims not kept: 45\%; likely-true ablation kept: 60\%; direct generation: 54\%), 67\% a panel prior below 50 (not kept: 54\%; ablation: 40\%; direct: 45\%), and 6\% both (not kept: 15\%). Within the pools, Sonnet's blind prior does not discriminate truth (AUC .51; panel prior .57; being kept .57). Five of 24 selection replies mention other models, and one says it did not estimate their ratings. The generator's in-context probabilities are not recorded in structured form, so we cannot test compliance with them directly.

\paragraph{Concentration.}
The harness's 240 hypotheses contain 93 distinct propositions (68 true); the stage-5 direct runs contain 62 (41), the stage-15 direct runs 76 (49) and the ablation 41 (33). Table~\ref{tab:conc} attributes the gain to propositions: a proposition's contribution is its summed $T-\pi$ in the harness arm minus that in the comparison arm, per run. The largest contribution in all three comparisons comes from a false claim with $\pi=.97$ that the comparison arms propose often and the harness avoids. Counting each distinct claim once per arm, the gain over direct generation is $+0.60$ per run (stage 5), $+0.57$ (stage 15) and $+0.71$ over the ablation, with 90\% run-bootstrap intervals above $+0.3$ in each case (the bootstrap understates the number of distinct claims, so these intervals are only indicative).

\begin{table}[h]
\centering\small
\setlength{\tabcolsep}{3pt}
\begin{tabular}{lcccccc}
\toprule
 & & \multicolumn{2}{c}{Contribution of top} & \multicolumn{3}{c}{$\Delta R$ after dropping the top $k$ from both arms} \\
\cmidrule(lr){3-4}\cmidrule(lr){5-7}
Comparison & $\Delta R$ & 5 & 10 & $k=3$ & $k=5$ & $k=10$ \\
\midrule
Direct (stage 5) & $+1.77$ & $+1.10$ & $+1.70$ & $+0.94$ ($<.001$) & $+0.67$ (.006) & $+0.07$ (.38) \\
New direct (stage 15) & $+1.95$ & $+0.98$ & $+1.47$ & $+1.20$ ($<.001$) & $+0.97$ (.001) & $+0.48$ (.02) \\
Likely-true ablation & $+1.96$ & $+1.35$ & $+1.98$ & $+0.90$ ($<.001$) & $+0.62$ (.005) & $-0.02$ (.53) \\
\bottomrule
\end{tabular}
\caption{Concentration of the stage-3 harness's gain over its three pre-registered comparison arms. Parentheses: one-sided permutation $p$ over runs. Dropping propositions chosen for their contribution biases these differences toward zero.}
\label{tab:conc}
\end{table}

\paragraph{Batch.}
The harness (stage 13) and the ablation (stage 14) generated their pools with the same prompt a day apart. The pools are similar: 29.5 and 29.3 distinct claims per run, precision .746 and .749, Other share .10 and .11, pool $R$ $+3.90$ and $+3.71$ per run (difference $+0.19$, two-sided $p=.76$). Their Other-type claims differ in precision (.72 against .46, on 69 and 76 claims). Decomposing the $+1.96$ difference between the arms: the rule beats random selection from its own pools by $+1.55$; random selections from the two batches differ by $+0.06$ (two-sided $p=.81$); and random selection beats likely-true selection in the ablation's pools by $+0.36$. Most of the difference therefore comes from the two selection steps, not from the batch, although the harness itself was not rerun alongside the ablation.

%% file: appendix_sensitivity.tex
\section{Sensitivity analyses}
\label{app:sens}
All analyses in this section are post hoc and use only cached runs and priors (script \texttt{post\_pos\_sensitivity.py}, 4,000 permutations, seed 28). Differences are in claims per run, arm minus comparison arm, 24 runs per arm; $^{*}$one-sided permutation $p<.05$, uncorrected.
\begin{table}[h]
\centering\scriptsize
\setlength{\tabcolsep}{2.5pt}
\begin{tabular}{llccccccccc}
\toprule
Bench. & Comparison & Def. & $\lambda{=}10$ & $\lambda{=}100$ & Prior & +Type & CV & Prop. & Raw & Sig./14 \\
\midrule
NB201 & Stage-3 harness vs direct & $+1.77$$^{*}$ & $+1.57$$^{*}$ & $+1.01$$^{*}$ & $+1.81$$^{*}$ & $+1.73$$^{*}$ & $+1.79$$^{*}$ & $+0.79$$^{*}$ & $+0.98$$^{*}$ & 14 \\
 & \quad vs new direct & $+1.95$$^{*}$ & $+1.77$$^{*}$ & $+1.36$$^{*}$ & $+2.00$$^{*}$ & $+1.89$$^{*}$ & $+1.97$$^{*}$ & $+0.97$$^{*}$ & $+1.36$$^{*}$ & 14 \\
 & \quad vs likely-true ablation & $+1.96$$^{*}$ & $+1.65$$^{*}$ & $+0.93$$^{*}$ & $+1.97$$^{*}$ & $+1.86$$^{*}$ & $+1.94$$^{*}$ & $+0.37$$^{*}$ & $+0.88$$^{*}$ & 14 \\
 & GPT measure vs direct & $+1.60$$^{*}$ & $+0.78$$^{*}$ & $-0.46$ & $+1.58$$^{*}$ & $+1.69$$^{*}$ & $+1.57$$^{*}$ & $-0.67$ & $-0.45$ & 12 \\
 & \quad vs pooled direct & $+2.24$$^{*}$ & $+1.50$$^{*}$ & $-0.15$ & $+2.34$$^{*}$ & $+2.34$$^{*}$ & $+2.20$$^{*}$ & $-0.58$ & $-0.16$ & 12 \\
 & \quad vs pooled + stage-3 rule & $+1.16$$^{*}$ & $+0.80$$^{*}$ & $-0.74$ & $+1.28$$^{*}$ & $+1.23$$^{*}$ & $+1.11$$^{*}$ & $-0.21$ & $-0.86$ & 12 \\
 & \quad vs 40-claim call & $+0.88$$^{*}$ & $+0.58$$^{*}$ & $-1.29$ & $+1.20$$^{*}$ & $+0.60$$^{*}$ & $+0.84$$^{*}$ & $-0.24$ & $-1.34$ & 12 \\
 & \quad vs shuffled values & $+1.40$$^{*}$ & $+1.85$$^{*}$ & $+2.31$$^{*}$ & $+1.44$$^{*}$ & $+1.45$$^{*}$ & $+1.40$$^{*}$ & $+1.85$$^{*}$ & $+2.23$$^{*}$ & 14 \\
 & \quad vs withheld values & $+0.89$$^{*}$ & $+0.10$ & $-0.77$ & $+0.83$$^{*}$ & $+1.05$$^{*}$ & $+0.88$$^{*}$ & $-1.21$ & $-0.72$ & 11 \\
HPO & GPT measure vs direct & $+0.94$$^{*}$ & $+0.63$$^{*}$ & $+0.50$$^{*}$ & $+0.94$$^{*}$ & $+0.48$$^{*}$ & $+0.97$$^{*}$ & $+0.62$$^{*}$ & $+0.56$$^{*}$ & 14 \\
 & \quad vs pooled direct & $+0.70$$^{*}$ & $+0.47$$^{*}$ & $+0.28$ & $+0.73$$^{*}$ & $+0.38$$^{*}$ & $+0.66$$^{*}$ & $+0.05$ & $+0.38$$^{*}$ & 12 \\
 & \quad vs pooled + stage-3 rule & $+0.94$$^{*}$ & $+0.83$$^{*}$ & $+1.52$$^{*}$ & $+0.20$ & $+0.21$ & $+0.98$$^{*}$ & $+0.52$$^{*}$ & $+1.13$$^{*}$ & 12 \\
 & \quad vs 40-claim call & $+0.09$ & $+0.12$ & $+0.11$ & $+0.16$ & $+0.05$ & $+0.04$ & $-0.16$ & $+0.11$ & 0 \\
 & \quad vs shuffled values & $+1.42$$^{*}$ & $+1.79$$^{*}$ & $+2.29$$^{*}$ & $+1.30$$^{*}$ & $+0.64$$^{*}$ & $+1.51$$^{*}$ & $+1.55$$^{*}$ & $+2.01$$^{*}$ & 14 \\
 & \quad vs withheld values & $+0.85$$^{*}$ & $+0.02$ & $-0.09$ & $+0.49$$^{*}$ & $+0.34$$^{*}$ & $+0.88$$^{*}$ & $+0.49$$^{*}$ & $0.00$ & 12 \\
NB201 & Opus measure vs direct & $+0.19$ & $-0.14$ & $-0.42$ & $+0.18$ & $-0.02$ & $+0.17$ & $-0.40$ & $-0.37$ & 0 \\
 & \quad vs withheld values & $+0.92$$^{*}$ & $+0.69$$^{*}$ & $+0.24$ & $+0.90$$^{*}$ & $+0.71$$^{*}$ & $+0.89$$^{*}$ & $-0.06$ & $+0.30$ & 12 \\
 & \quad vs shuffled values & $+1.84$$^{*}$ & $+1.85$$^{*}$ & $+1.74$$^{*}$ & $+1.67$$^{*}$ & $+1.63$$^{*}$ & $+1.85$$^{*}$ & $+1.02$$^{*}$ & $+1.65$$^{*}$ & 14 \\
HPO & Opus measure vs direct & $+0.42$ & $+0.21$ & $+0.10$ & $+0.42$ & $+0.47$$^{*}$ & $+0.38$ & $+0.02$ & $+0.20$ & 1 \\
NB201 & Sonnet measure vs direct & $-0.57$ & $-1.07$ & $-2.07$ & $-0.65$ & $-0.74$ & $-0.57$ & $-0.81$ & $-2.11$ & 0 \\
HPO & Sonnet measure vs direct & $+0.11$ & $-0.68$ & $-0.33$ & $-0.62$ & $-0.19$ & $+0.08$ & $+0.18$ & $-0.49$ & 0 \\
\bottomrule
\end{tabular}
\caption{$R$ under alternative calibrations, scored against the leave-one-out Claude panel. Def.: the calibration of the main text (ridge $\lambda=10^{-3}$, prior clipped to $[0.5,99.5]$, covariates logit prior, order, direction and logit$\times$order). $\lambda$: ridge penalty. Prior: logit prior only. +Type: adds indicators for the edited value and the condition (NB201: operations; HPO: band, edit direction and edited parameter). CV: 10-fold cross-fitting by cell. Prop.: calibration fitted only on the claims the two compared arms proposed, not on the whole space; with an intercept this makes the two arms' summed $R$ zero, so it asks only whether one arm's claims beat the other's at equal prior. Raw: uncalibrated $\sum(T-p/100)$. Sig./14: number of the 14 calibrations (the columns shown plus $\lambda\in\{0,0.1,1\}$ and clipping at 0.1, 1, 2.5 and 5) with $p<.05$; Raw is not among them. Clipping never changes a difference by more than 0.02, cross-fitting by more than 0.09.}
\label{tab:sens_cal}
\end{table}
\begin{table}[h]
\centering\scriptsize
\setlength{\tabcolsep}{2.5pt}
\begin{tabular}{llccccccccc}
\toprule
Bench. & Comparison & Def. & $\lambda{=}10$ & $\lambda{=}100$ & Prior & +Type & CV & Prop. & Raw & Sig./14 \\
\midrule
NB201 & Stage-3 harness vs direct & $+1.25$$^{*}$ & $+1.19$$^{*}$ & $+1.00$$^{*}$ & $+1.29$$^{*}$ & $+1.14$$^{*}$ & $+1.26$$^{*}$ & $+0.84$$^{*}$ & $+1.16$$^{*}$ & 14 \\
 & \quad vs new direct & $+1.55$$^{*}$ & $+1.50$$^{*}$ & $+1.35$$^{*}$ & $+1.60$$^{*}$ & $+1.44$$^{*}$ & $+1.57$$^{*}$ & $+1.11$$^{*}$ & $+1.48$$^{*}$ & 14 \\
 & \quad vs likely-true ablation & $+1.16$$^{*}$ & $+1.08$$^{*}$ & $+0.85$$^{*}$ & $+1.19$$^{*}$ & $+1.02$$^{*}$ & $+1.17$$^{*}$ & $+0.40$$^{*}$ & $+1.05$$^{*}$ & 14 \\
 & GPT measure vs direct & $+2.69$$^{*}$ & $+2.28$$^{*}$ & $+1.04$$^{*}$ & $+2.46$$^{*}$ & $+2.61$$^{*}$ & $+2.69$$^{*}$ & $+0.49$ & $+0.29$ & 13 \\
 & \quad vs pooled direct & $+2.97$$^{*}$ & $+2.66$$^{*}$ & $+1.34$$^{*}$ & $+2.96$$^{*}$ & $+3.06$$^{*}$ & $+2.96$$^{*}$ & $+0.39$ & $+0.49$ & 13 \\
 & \quad vs pooled + stage-3 rule & $+2.02$$^{*}$ & $+1.92$$^{*}$ & $+0.81$$^{*}$ & $+2.00$$^{*}$ & $+2.11$$^{*}$ & $+1.98$$^{*}$ & $+0.06$ & $-0.09$ & 13 \\
 & \quad vs 40-claim call & $+1.28$$^{*}$ & $+1.27$$^{*}$ & $+0.09$ & $+1.61$$^{*}$ & $+1.20$$^{*}$ & $+1.25$$^{*}$ & $-0.07$ & $-0.79$ & 12 \\
 & \quad vs shuffled values & $+0.98$$^{*}$ & $+1.10$$^{*}$ & $+1.36$$^{*}$ & $+1.03$$^{*}$ & $+0.95$$^{*}$ & $+0.98$$^{*}$ & $+0.89$$^{*}$ & $+0.69$$^{*}$ & 14 \\
 & \quad vs withheld values & $+1.89$$^{*}$ & $+1.47$$^{*}$ & $+0.46$ & $+1.62$$^{*}$ & $+1.87$$^{*}$ & $+1.93$$^{*}$ & $+0.03$ & $-0.07$ & 12 \\
HPO & GPT measure vs direct & $+0.61$$^{*}$ & $+0.50$$^{*}$ & $+0.41$$^{*}$ & $+0.61$$^{*}$ & $+0.24$ & $+0.64$$^{*}$ & $+0.39$$^{*}$ & $+0.45$$^{*}$ & 13 \\
 & \quad vs pooled direct & $+0.72$$^{*}$ & $+0.51$$^{*}$ & $+0.28$ & $+0.72$$^{*}$ & $+0.60$$^{*}$ & $+0.71$$^{*}$ & $+0.02$ & $+0.29$$^{*}$ & 12 \\
 & \quad vs pooled + stage-3 rule & $+0.53$$^{*}$ & $+0.47$$^{*}$ & $+1.06$$^{*}$ & $+0.39$$^{*}$ & $+0.23$ & $+0.54$$^{*}$ & $+0.48$$^{*}$ & $+0.70$$^{*}$ & 13 \\
 & \quad vs 40-claim call & $+0.42$$^{*}$ & $+0.27$ & $+0.14$ & $+0.45$$^{*}$ & $+0.56$$^{*}$ & $+0.40$$^{*}$ & $-0.20$ & $+0.12$ & 11 \\
 & \quad vs shuffled values & $+1.01$$^{*}$ & $+1.32$$^{*}$ & $+1.98$$^{*}$ & $+0.94$$^{*}$ & $+0.58$$^{*}$ & $+1.04$$^{*}$ & $+1.02$$^{*}$ & $+1.69$$^{*}$ & 14 \\
 & \quad vs withheld values & $+0.51$$^{*}$ & $+0.14$ & $-0.01$ & $+0.44$$^{*}$ & $+0.14$ & $+0.53$$^{*}$ & $+0.37$$^{*}$ & $+0.24$ & 10 \\
NB201 & Opus measure vs direct & $+0.91$$^{*}$ & $+0.58$ & $+0.04$ & $+0.67$$^{*}$ & $+0.72$$^{*}$ & $+0.88$$^{*}$ & $+0.02$ & $-0.12$ & 11 \\
 & \quad vs withheld values & $+1.20$$^{*}$ & $+1.07$$^{*}$ & $+0.56$$^{*}$ & $+1.24$$^{*}$ & $+1.03$$^{*}$ & $+1.19$$^{*}$ & $+0.24$ & $+0.51$ & 13 \\
 & \quad vs shuffled values & $+1.77$$^{*}$ & $+1.81$$^{*}$ & $+1.74$$^{*}$ & $+1.79$$^{*}$ & $+1.65$$^{*}$ & $+1.76$$^{*}$ & $+1.24$$^{*}$ & $+1.64$$^{*}$ & 14 \\
HPO & Opus measure vs direct & $+0.48$$^{*}$ & $+0.29$ & $+0.10$ & $+0.47$$^{*}$ & $+0.57$$^{*}$ & $+0.49$$^{*}$ & $-0.02$ & $+0.18$ & 11 \\
NB201 & Sonnet measure vs direct & $-2.05$ & $-2.12$ & $-2.33$ & $-1.90$ & $-2.19$ & $-2.06$ & $-2.05$ & $-2.03$ & 0 \\
HPO & Sonnet measure vs direct & $+0.25$ & $-0.35$ & $-0.23$ & $-0.30$ & $-0.14$ & $+0.31$ & $+0.50$$^{*}$ & $-0.32$ & 1 \\
\bottomrule
\end{tabular}
\caption{$R$ under alternative calibrations, scored against each generator's own blind three-way ratings. Def.: the calibration of the main text (ridge $\lambda=10^{-3}$, prior clipped to $[0.5,99.5]$, covariates logit prior, order, direction and logit$\times$order). $\lambda$: ridge penalty. Prior: logit prior only. +Type: adds indicators for the edited value and the condition (NB201: operations; HPO: band, edit direction and edited parameter). CV: 10-fold cross-fitting by cell. Prop.: calibration fitted only on the claims the two compared arms proposed, not on the whole space; with an intercept this makes the two arms' summed $R$ zero, so it asks only whether one arm's claims beat the other's at equal prior. Raw: uncalibrated $\sum(T-p/100)$. Sig./14: number of the 14 calibrations (the columns shown plus $\lambda\in\{0,0.1,1\}$ and clipping at 0.1, 1, 2.5 and 5) with $p<.05$; Raw is not among them. Clipping never changes a difference by more than 0.02, cross-fitting by more than 0.09.}
\label{tab:sens_cal_own}
\end{table}
\begin{table}[h]
\centering\scriptsize
\setlength{\tabcolsep}{2.5pt}
\begin{tabular}{llcccccc}
\toprule
 & & & \multicolumn{2}{c}{Stage-3 harness (Sonnet)} & \multicolumn{2}{c}{GPT measure-first} & Judge-opt. \\
\cmidrule(lr){4-5}\cmidrule(lr){6-7}
Bench. & Truth rule & True (rate) & $\Delta R$ vs direct / new direct & Prec. & $\Delta R$ LOO / own & Prec. & bound met \\
\midrule
NB201 & all, $\tau=1$ & 419 (.241) & $+1.08$$^{*}$ / $+1.44$$^{*}$ & .90$\to$.88 & $+1.06$$^{*}$ / $+1.30$$^{*}$ & .85$\to$.62$^{\times}$ & 0/3 \\
 & all, $\tau=1.5$ & 288 (.166) & $+1.14$$^{*}$ / $+1.51$$^{*}$ & .89$\to$.86 & $+1.56$$^{*}$ / $+2.19$$^{*}$ & .79$\to$.58$^{\times}$ & 0/3 \\
 & all, $\tau=2$ (main) & 198 (.114) & $+1.77$$^{*}$ / $+1.95$$^{*}$ & .79$\to$.83 & $+1.60$$^{*}$ / $+2.69$$^{*}$ & .69$\to$.50$^{\times}$ & 0/3 \\
 & all, $\tau=3$$^{\ddagger}$ & 108 (.062) & $+1.59$$^{*}$ / $+1.84$$^{*}$ & .59$\to$.63 & $+1.34$$^{*}$ / $+2.58$$^{*}$ & .49$\to$.32$^{\times}$ & 0/3 \\
 & 2 of 3, $\tau=2$ & 339 (.195) & $+1.63$$^{*}$ / $+1.78$$^{*}$ & .86$\to$.90 & $+1.98$$^{*}$ / $+2.51$$^{*}$ & .76$\to$.62 & 0/3 \\
 & mean, $\tau=2$ & 314 (.180) & $+1.36$$^{*}$ / $+1.68$$^{*}$ & .91$\to$.91 & $+1.56$$^{*}$ / $+2.14$$^{*}$ & .78$\to$.58$^{\times}$ & 0/3 \\
HPO & all, $\tau=0.25$ & 52 (.176) & -- & -- & $+0.86$$^{*}$ / $+0.48$$^{*}$ & .94$\to$.98 & 0/3 \\
 & all, $\tau=0.5$ (main) & 41 (.139) & -- & -- & $+0.94$$^{*}$ / $+0.61$$^{*}$ & .90$\to$.95 & 0/3 \\
 & all, $\tau=1$$^{\ddagger}$ & 22 (.074) & -- & -- & $+0.75$$^{*}$ / $+0.51$$^{*}$ & .71$\to$.75 & 0/3 \\
 & 2 of 3, $\tau=0.5$ & 84 (.284) & -- & -- & $+0.61$$^{*}$ / $+0.20$$^{*}$ & .98$\to$.99 & 0/3 \\
 & mean, $\tau=0.5$ & 88 (.297) & -- & -- & $+0.56$$^{*}$ / $+0.20$$^{*}$ & .98$\to$1.00 & 0/3 \\
\bottomrule
\end{tabular}
\caption{Results under alternative truth definitions. A claim is true when the conditional difference exceeds $\tau$ pp with the stated sign on all three datasets (main text), on at least two of three, or in the mean over datasets; $R$ is recalibrated under each definition. Stage-3 harness: Sonnet on NB201 against the stage-5 direct runs and the new stage-15 direct runs (leave-one-out Claude panel). Not shown as a column: the harness's advantage over the likely-true ablation in the six NB201 rows is $+0.69$, $+1.12$, $+1.96$, $+1.91$, $+1.73$ and $+1.35$ (all $p\le.002$), with the precision bound met in all six. GPT measure-first against direct generation under the leave-one-out Claude panel and GPT's own blind ratings; $^{\times}$precision below 0.8 times direct generation's. Judge-opt.: number of the three generators for which the judge-optimized harness keeps precision at least 0.8 times direct generation's. $^{\ddagger}$At these two thresholds the calibration's undamped Newton fit diverges (every $\pi$ goes to 1); these rows use the same model fitted by Newton's method with step halving on the penalized likelihood (script \texttt{post\_pos\_round6d.py}), which gives the same fit as the undamped one for every other truth definition here and for all 1,657 fits of the calibration and panel analyses (\texttt{post\_pos\_round6e.py}).}
\label{tab:sens_truth}
\end{table}
\begin{table}[h]
\centering\scriptsize
\setlength{\tabcolsep}{2.5pt}
\begin{tabular}{llccc}
\toprule
Bench. & Comparison & One member & Two members & Three members \\
\midrule
NB201 & Stage-3 harness (Sonnet) & H $+1.26$$^{*}$, O $+1.51$$^{*}$, G $+1.82$$^{*}$ & H+O $+1.77$$^{*}$, H+G $+1.74$$^{*}$, O+G $+1.71$$^{*}$ & H+O+G $+1.78$$^{*}$ \\
 & GPT measure & H $-0.56$, S $-0.12$, O $+1.53$$^{*}$ & H+S $+0.30$, H+O $+1.43$$^{*}$, S+O $+1.71$$^{*}$ & H+S+O $+1.60$$^{*}$ \\
 & Opus measure & H $-0.31$, S $-0.21$, G $+1.25$$^{*}$ & H+S $+0.19$, H+G $+0.91$$^{*}$, S+G $+0.84$$^{*}$ & H+S+G $+0.90$$^{*}$ \\
 & Sonnet measure & H $-1.97$, O $-0.90$, G $-0.17$ & H+O $-0.57$, H+G $-0.30$, O+G $-0.23$ & H+O+G $-0.10$ \\
HPO & GPT measure & H $+0.42$, S $+0.86$$^{*}$, O $+0.78$$^{*}$ & H+S $+0.74$$^{*}$, H+O $+0.72$$^{*}$, S+O $+1.01$$^{*}$ & H+S+O $+0.94$$^{*}$ \\
 & Opus measure & H $+0.04$, S $+0.54$$^{*}$, G $+0.40$$^{*}$ & H+S $+0.42$, H+G $+0.34$, S+G $+0.60$$^{*}$ & H+S+G $+0.52$$^{*}$ \\
 & Sonnet measure & H $-0.52$, O $0.00$, G $-0.15$ & H+O $+0.11$, H+G $-0.07$, O+G $+0.07$ & H+O+G $+0.05$ \\
\bottomrule
\end{tabular}
\caption{$\Delta R$ against direct generation under every panel drawn from the models other than the generator, grouped by panel size (H/S/O/G: Haiku/Sonnet/Opus/GPT; the panel is the mean of its members' three-way priors, recalibrated per panel). The leave-one-out Claude panel of the main text is H+O for Sonnet, H+S for Opus and H+S+O for GPT.}
\label{tab:sens_panel}
\end{table}

%% file: appendix_definitions.tex
\section{Truth definition and harness procedure}
\label{app:defs}

\paragraph{Truth.}
Let $\mathcal{X}$ be the benchmark's configurations and $\mathrm{acc}_D(x)$ the validation accuracy of $x$ on dataset $D$, in percentage points (NB201: CIFAR-10, CIFAR-100 and ImageNet16-120; HPOBench: the mean over five training seeds on each of three OpenML tasks). An edit $e$ maps a parent $x$ to a child $e(x)$ that differs from $x$ in one parameter $j$: on NAS-Bench-201, edge $j$ is set to operation $b$; on HPOBench, parameter $j$ moves one grid step up or down. The edit's parents $P_e$ are the configurations for which it changes something: on NAS-Bench-201 those with $x_j\neq b$, so no parent is a no-op; on HPOBench those not at the end of the grid in the direction of the step. Its effect on a parent is $\Delta_D(x)=\mathrm{acc}_D(e(x))-\mathrm{acc}_D(x)$. A condition $(k,c)$ splits the parents into $S=\{x\in P_e: x_k=c\}$ (on HPOBench, $x_k$ lies in band $c$) and $P_e\setminus S$, and the conditional difference is the unweighted difference of means
\[
d_D(e,k,c)=\frac{1}{|S|}\sum_{x\in S}\Delta_D(x)\;-\;\frac{1}{|P_e\setminus S|}\sum_{x\in P_e\setminus S}\Delta_D(x).
\]
A claim $(e,k,c,s)$ with direction $s=+1$ (higher) or $-1$ (lower) is true when $s\,d_D(e,k,c)\geq\tau$ on every dataset $D$; the variants of Appendix~\ref{app:sens} replace ``every dataset'' with ``at least two of three'' or with the mean of $d_D$ over datasets. The condition may be on the edited parameter itself ($k=j$), where it states which value the edit replaces. A cell is kept only when both groups have at least 100 parents (NAS-Bench-201) or 500 (HPOBench).

\emph{Counts.} NAS-Bench-201 has $5^6=15{,}625$ architectures. Its 30 edits (six edges, five operations) each have $4\cdot5^5=12{,}500$ parents, and 29 conditions (six edges times five operations, less $k=j$, $c=b$, which no parent satisfies). All $30\times29=870$ cells meet the minimum, giving 1,740 directional claims. HPOBench has five parameters with ten grid levels each, except depth with three. Depth's levels are its bands; for the others, levels 1--3 are low, 4--7 mid and 8--10 high. Ten edits (five parameters, up or down) times 15 conditions (five parameters, three bands) give 150 cells, of which 148 meet the minimum (296 claims).

\emph{Equivalent wordings.} On HPOBench, for $k\neq j$, the pairs (parent, child) of ``move $j$ up'' are the pairs (child, parent) of ``move $j$ down'', and the edit does not change $x_k$. So ``moving $j$ up has a higher effect when $x_k$ is in band $c$'' and ``moving $j$ down has a lower effect when $x_k$ is in band $c$'' have the same $d_D$ up to sign and the same truth. They are merged into one claim before scoring, and their priors are averaged.

\emph{Worked example.} Take the claim ``setting edge 2 to a $3{\times}3$ convolution has a lower effect when edge 0 is \texttt{none}''. Of the edit's 12,500 parents, 2,500 have edge 0 set to \texttt{none}. The conditional differences are $-1.93$, $-2.73$ and $-2.57$ pp on CIFAR-10, CIFAR-100 and ImageNet16-120, so the stated direction holds on all three datasets. The CIFAR-10 magnitude is below $\tau=2$, so the claim is false in the main analysis, though true at $\tau=1$ and $1.5$. Direct generation proposes this claim often, the calibrated panel gives it $\pi=.97$, and it is the largest single contributor to the stage-3 harness's gain (Section~\ref{sec:sel}; Appendix~\ref{app:decomp} counts such near misses).

\paragraph{Optimizer.}
The optimizer is a single Opus prompt per round. It sees the harness fields and their allowed values, the token cap, and the definition of the score. For the \SV{} arms the score is ``the number of the 10 hypotheses that turn out TRUE when tested against measured results AND that other strong language models rate as less than 50\% likely to hold''. It also sees the full history of harnesses tried, with mean score, mean tokens and status, the current best harness, and one unlabeled output of it. It does not see per-hypothesis truth or priors. It proposes 3 candidates per round for 3 rounds. Each candidate is run 6 times, and the round's best is accepted if it beats the incumbent. Text fields (a selection rule of at most 50 words, a strategy of at most 150) must be domain-general: a regular expression rejects benchmark-specific terms.

\paragraph{The stage-2 to stage-3 fix.}
In stage 2 that expression's pattern \texttt{conv\textbackslash w*} also matched ``conventional''. It wrongly rejected two of the three round-1 candidates of the \SV{} arm, and one round-2 candidate of the true-count arm. Stage 3 narrowed the pattern to \emph{conv}, \emph{convolution\ldots} and \emph{conv} followed by a digit, and reran the \SV{} arm with fresh optimizer calls. It also re-estimated the starting harness (direct generation) on 12 runs instead of 6: the 6-run estimate, 1.5, had been above stage-1 direct generation's 1.25, which made candidates harder to accept. From the 12-run start of 1.42, one of the nine stage-3 candidates was accepted: the stage-3 harness, in round 3, with a training score of 1.67.

\paragraph{Where the thresholds come from.}
The selection rule, including its 55\% and 50\% thresholds, was written by the optimizer in its round-3 proposal, as free text in the selection-rule field. We did not choose or tune either value. The 50\% echoes the score definition the optimizer was shown. Its stated rationale was: ``Selects directly on the gap between the generator's calibrated belief and naive expectation, which is exactly what the scoring rule rewards, without pushing for risky sign reversals.''

\paragraph{How the harness runs.}
A run makes two stateless calls to the generator.

\emph{Generation call.} The prompt contains the benchmark description, an attempt number, ``Approach to follow:'' followed by the strategy (Appendix~\ref{app:prompts}), ``Propose 30 conditional hypotheses about this benchmark that you believe are true'', and the output format. The format has fields for the edit, the condition, the direction and a one-sentence ``why''. It has no field for a probability. None of the 709 hypotheses in the 24 confirmation pools states a probability, although the strategy asks for one.

\emph{Selection call.} This is a separate call. It contains the benchmark description, the 30 candidates as JSON (with their ``why''), and ``Select the 10 that best satisfy this criterion:'' followed by the rule, ``Copy them exactly'', and the format. The ablation's selection call reads ``Select the 10 most likely to be true'' instead. Any probabilities the rule refers to therefore exist only inside the selection call and are not recorded.

One selection reply put this directly: ``The probabilities come from my own reasoning about NAS-Bench-201, and I did not estimate how other models would rate them.'' This is why compliance (Appendix~\ref{app:selection}) can only be checked against separately elicited ratings.

%% file: appendix_decomp.tex
\section{Surprising truths or avoided errors, and the precision bound}
\label{app:decomp}
All analyses in this section are post hoc and use cached runs only. The script is \texttt{post\_pos\_decomp.py}.

\paragraph{Splitting $R$.}
Because $T_h\in\{0,1\}$, a run's score splits as
\[
R=\underbrace{\textstyle\sum_{h\,\text{true}}(1-\pi_h)}_{\text{credit}}\;-\;\underbrace{\textstyle\sum_{h\,\text{false}}\pi_h}_{\text{cost}}.
\]
A difference in $R$ can therefore come from true claims, each credited by how little the panel expected it (credit), or from fewer false claims the panel expected (cost avoided). Only the first is surprise in the intended sense, although it also gives partial credit to true claims the panel expected. The second rewards avoiding consensus errors, as a precision gain does. Table~\ref{tab:decomp} splits every headline comparison.

\emph{Stage-3 harness.} Against each of its three comparison arms, the credit term alone is significant:
\begin{itemize}
\item $+1.04$ of $+1.77$ against direct generation;
\item $+1.13$ of $+1.95$ against the new stage-15 direct runs;
\item $+1.22$ of $+1.96$ against the ablation;
\item $+0.66$ to $+1.09$ under Sonnet's own ratings.
\end{itemize}
All are $p\leq.001$. Its true claims have a lower mean calibrated prior than direct generation's (.58 against .70), while its precision is higher.

The rest of the gain, $+0.73$ to $+0.82$, is cost avoided. Most of that is near misses: claims whose stated direction holds on all three datasets but by less than $\tau=2$ pp. Direct generation pays 0.97 per run for near misses and the harness 0.44. The largest is the worked example of Appendix~\ref{app:defs}. Against the ablation, all of the cost avoided is near misses ($+0.78$ for near misses, $-0.04$ for other false claims). The cost term is thus partly an artifact of the threshold: at $\tau=1$ and 1.5 pp, where most near misses become true, almost all of the gain is credit, and at $\tau=3$ credit and cost avoided are about equal (Table~\ref{tab:crossed}). The credit term is the part of the gain that is surprise in the intended sense.

\emph{GPT measure-first.} On HPOBench the gain is mostly credit:
\begin{itemize}
\item $+0.69$ of $+0.94$ against direct generation;
\item $+0.76$ against pooled direct generation, where cost avoided is $-0.06$ and the total $+0.70$.
\end{itemize}
On NAS-Bench-201 it is mostly cost avoided ($+1.14$ of $+1.60$ against direct generation), and against withheld values the credit term is $-0.12$. This is consistent with the NB201 gain failing the precision bound: its false claims are many (5.0 per run against 3.1) but have low priors.

\emph{Judge-optimized harness.} Where it raised $R$, half or more of the rise is cost avoided:
\begin{itemize}
\item Sonnet on NB201: $+0.43$ credit, $+0.45$ cost avoided;
\item GPT on NB201: $-0.38$ credit, $+1.41$ cost avoided.
\end{itemize}
Its many false claims are ones the panel also doubted, so they cost little. This is the case the precision bound exists for.

\paragraph{The precision bound.}
The bound of 0.8 times the comparison arm's precision was first pre-registered in stage 2, before that stage's runs, and is part of the measurement criteria of stages 4--6 and 8--10, the stage-12 rescoring and the stage-13 harness confirmation; stages 13c, 14 and 15 did not include it (Appendix~\ref{app:checks}). It allows a run to give up about one in five true claims in exchange for surprise. Without such a bound, $R$ can rise when precision collapses (above). Table~\ref{tab:bound} gives each comparison's precision ratio and the verdict at bounds from 0.6 to 1.0.
\begin{itemize}
\item \emph{Stage-3 harness.} It meets every bound up to 1.0 (ratios 1.00 to 1.12).
\item \emph{Judge-optimized harness.} It fails every bound from 0.6 up (ratios .32 to .54).
\item \emph{GPT measure-first on NB201.} The ratios are .61 to .72. It would pass against direct, pooled and withheld-value generation at a bound of 0.7, and fails at 0.8.
\item \emph{GPT measure-first on HPOBench.} The ratios are .97 to 1.15. It passes every bound up to 0.9, and at 1.0 fails against pooled direct generation and the 40-claim call.
\end{itemize}
The verdicts on the stage-3 harness and the judge-optimized harness therefore do not depend on the bound. The measurement verdicts do: GPT's NB201 gain would pass at 0.7, Opus's NB201 comparison with direct generation (ratio .86) would fail at 0.9, and two of GPT's HPOBench comparisons would fail at 1.0.

\paragraph{Resampling propositions as well as runs (script \texttt{post\_pos\_round6a.py}, with \texttt{post\_pos\_round6c.py} and \texttt{post\_pos\_round6f.py}).}
The harness's 240 hypotheses contain 93 distinct propositions, and a few recurring ones carry much of the gain (Appendix~\ref{app:selection}). The run-level intervals of the main text treat these propositions as fixed. Table~\ref{tab:crossed} also treats them as a sample. In each of 4,000 draws, the runs of each arm are resampled, the union of the two arms' distinct propositions (106 to 125) is resampled with replacement, and each run's $R$, credit and cost are recomputed with every proposition weighted by its resampled count. Under the main panel, the gain stays above zero against all three arms at the main threshold ($p_\times=.015$, .004, .024) and at every other truth definition except $\tau=1$ against the ablation ($p_\times=.102$); at $\tau=3$ it is just above zero ($p_\times=.024$ to $.049$). With GPT added to the panel (main threshold) it is $p_\times\leq.009$. Against Sonnet's own blind prior it is not ($p_\times=.062$ to $.192$). The credit term is more fragile than the total: at the main threshold $p_\times=.054$ against direct generation and .028 and .045 against the other arms, and it is not significant against Sonnet's own prior or at $\tau=3$; at $\tau=1$, 1.5, and the two- and mean-agreement rules it is ($p_\times\leq.013$). The positive result that survives this analysis is that the selection step raises $R$ relative to the Claude panels on propositions beyond those sampled; how much of that is credit for true low-prior claims, rather than fewer expected errors, depends on the threshold and the panel.

\input{appendix_decomp_tables}
\input{appendix_round6_tables}

%% file: appendix_decomp_tables.tex
\begin{table}[h]
\centering\scriptsize
\setlength{\tabcolsep}{3pt}
\begin{tabular}{llcccccccc}
\toprule
 & & \multicolumn{4}{c}{Leave-one-out Claude panel} & \multicolumn{3}{c}{Generator's own ratings} & \\
\cmidrule(lr){3-6}\cmidrule(lr){7-9}
Bench. & Comparison & $\Delta R$ & Credit & Cost avoided & $\bar\pi$ true & $\Delta R$ & Credit & Cost avoided & Prec.\ ratio \\
\midrule
NB201 & Stage-3 harness vs direct & $+1.77$ & $+1.04$$^{*}$ & $+0.73$$^{*}$ & 0.58/0.70 & $+1.25$ & $+0.80$$^{*}$ & $+0.45$$^{*}$ & 1.05 \\
 & \quad vs new direct & $+1.95$ & $+1.13$$^{*}$ & $+0.82$$^{*}$ & 0.58/0.69 & $+1.55$ & $+1.09$$^{*}$ & $+0.46$$^{*}$ & 1.12 \\
 & \quad vs likely-true ablation & $+1.96$ & $+1.22$$^{*}$ & $+0.74$$^{*}$ & 0.58/0.73 & $+1.16$ & $+0.66$$^{*}$ & $+0.50$$^{*}$ & 1.00 \\
HPO & GPT measure-first vs direct & $+0.94$ & $+0.69$$^{*}$ & $+0.25$$^{*}$ & 0.69/0.77 & $+0.61$ & $+0.40$$^{*}$ & $+0.21$ & 1.05 \\
 & \quad vs pooled direct & $+0.70$ & $+0.76$$^{*}$ & $-0.06$ & 0.69/0.78 & $+0.72$ & $+0.80$$^{*}$ & $-0.09$ & 0.98 \\
 & \quad vs pooled + stage-3 rule & $+0.94$ & $+0.69$$^{*}$ & $+0.25$$^{*}$ & 0.69/0.65 & $+0.53$ & $+0.30$$^{*}$ & $+0.24$$^{*}$ & 1.15 \\
 & \quad vs 40-claim call & $+0.09$ & $+0.18$ & $-0.09$ & 0.69/0.71 & $+0.42$ & $+0.57$$^{*}$ & $-0.15$ & 0.97 \\
 & \quad vs shuffled values & $+1.42$ & $+0.93$$^{*}$ & $+0.49$$^{*}$ & 0.69/0.70 & $+1.01$ & $+0.56$$^{*}$ & $+0.45$$^{*}$ & 1.63 \\
 & \quad vs withheld values & $+0.85$ & $+0.53$$^{*}$ & $+0.32$$^{*}$ & 0.69/0.78 & $+0.51$ & $+0.20$ & $+0.31$$^{*}$ & 1.05 \\
NB201 & GPT measure-first vs direct & $+1.60$ & $+0.46$$^{*}$ & $+1.14$$^{*}$ & 0.44/0.68 & $+2.69$ & $+1.41$$^{*}$ & $+1.28$$^{*}$ & 0.72 \\
 & \quad vs pooled direct & $+2.24$ & $+0.97$$^{*}$ & $+1.28$$^{*}$ & 0.44/0.75 & $+2.97$ & $+1.50$$^{*}$ & $+1.47$$^{*}$ & 0.71 \\
 & \quad vs shuffled values & $+1.40$ & $+1.66$$^{*}$ & $-0.26$$^{*}$ & 0.44/0.44 & $+0.98$ & $+1.48$$^{*}$ & $-0.50$$^{*}$ & 2.64 \\
 & \quad vs withheld values & $+0.89$ & $-0.12$ & $+1.01$$^{*}$ & 0.44/0.59 & $+1.89$ & $+0.84$$^{*}$ & $+1.05$$^{*}$ & 0.72 \\
 & Judge-opt.\ vs direct, Sonnet & $+0.88$ & $+0.43$$^{*}$ & $+0.45$$^{*}$ & 0.35/0.70 & $-1.05$ & $-1.13$$^{*}$ & $+0.08$ & 0.54 \\
 & \quad Opus & $-2.24$ & $-2.37$$^{*}$ & $+0.12$ & 0.25/0.48 & $-1.07$ & $-0.74$$^{*}$ & $-0.32$$^{*}$ & 0.32 \\
 & \quad GPT & $+1.04$ & $-0.38$ & $+1.41$$^{*}$ & 0.30/0.68 & $+1.91$ & $+0.57$$^{*}$ & $+1.35$$^{*}$ & 0.38 \\
HPO & Judge-opt.\ vs direct, Sonnet & $-0.89$ & $+0.11$ & $-1.00$$^{*}$ & 0.60/0.79 & $-1.63$ & $-0.64$$^{*}$ & $-0.99$$^{*}$ & 0.53 \\
 & \quad Opus & $-1.69$ & $-0.74$$^{*}$ & $-0.95$$^{*}$ & 0.36/0.60 & $-1.24$ & $+0.29$$^{*}$ & $-1.53$$^{*}$ & 0.47 \\
 & \quad GPT & $-0.87$ & $-0.07$ & $-0.80$$^{*}$ & 0.59/0.77 & $-0.58$ & $-0.02$ & $-0.56$$^{*}$ & 0.39 \\
\bottomrule
\end{tabular}
\caption{$R$ split into the credit for true claims, $\sum_{\text{true}}(1-\pi)$, and the cost of false claims, $\sum_{\text{false}}\pi$, per run; $\Delta R$ is the difference in credit plus the difference in cost avoided (post hoc, paper-default calibration). $^{*}$one-sided permutation $p<.05$ in the direction shown. $\bar\pi$ true: mean calibrated prior of the true claims in the first and the second arm. Prec.\ ratio: precision of the first arm over the second; the pre-registered bound is 0.8. For the stage-3 harness, near misses (claims whose stated direction holds on all three datasets but by less than $\tau=2$ pp) account for 0.44 of its false-claim cost per run against 0.97 for direct generation, 0.94 for the new stage-15 direct runs and 1.22 for the ablation; its other false claims cost 0.10, against 0.31, 0.42 and 0.07.}
\label{tab:decomp}
\end{table}

\begin{table}[h]
\centering\scriptsize
\begin{tabular}{llcccccccc}
\toprule
Bench. & Comparison & Prec. & Prec.\ (comparison) & Ratio & 0.6 & 0.7 & 0.8 & 0.9 & 1.0 \\
\midrule
NB201 & Stage-3 harness vs direct & 0.83 & 0.79 & 1.05 & yes & yes & yes & yes & yes \\
 & \quad vs new direct & 0.83 & 0.74 & 1.12 & yes & yes & yes & yes & yes \\
 & \quad vs likely-true ablation & 0.83 & 0.82 & 1.00 & yes & yes & yes & yes & yes \\
 & GPT measure-first vs direct & 0.50 & 0.69 & 0.72 & yes & yes & -- & -- & -- \\
 & \quad vs pooled direct & 0.50 & 0.70 & 0.71 & yes & yes & -- & -- & -- \\
 & \quad vs pooled + stage-3 rule & 0.50 & 0.73 & 0.68 & yes & -- & -- & -- & -- \\
 & \quad vs 40-claim call & 0.50 & 0.82 & 0.61 & yes & -- & -- & -- & -- \\
 & \quad vs withheld values & 0.50 & 0.69 & 0.72 & yes & yes & -- & -- & -- \\
 & Opus measure-first vs direct & 0.75 & 0.87 & 0.86 & yes & yes & yes & -- & -- \\
HPO & GPT measure-first vs direct & 0.95 & 0.90 & 1.05 & yes & yes & yes & yes & yes \\
 & \quad vs pooled direct & 0.95 & 0.96 & 0.98 & yes & yes & yes & yes & -- \\
 & \quad vs pooled + stage-3 rule & 0.95 & 0.82 & 1.15 & yes & yes & yes & yes & yes \\
 & \quad vs 40-claim call & 0.95 & 0.97 & 0.97 & yes & yes & yes & yes & -- \\
 & \quad vs withheld values & 0.95 & 0.90 & 1.05 & yes & yes & yes & yes & yes \\
 & Opus measure-first vs direct & 0.89 & 0.90 & 0.99 & yes & yes & yes & yes & -- \\
NB201 & Judge-opt.\ vs direct, Sonnet & 0.43 & 0.79 & 0.54 & -- & -- & -- & -- & -- \\
 & Judge-opt.\ vs direct, Opus & 0.28 & 0.87 & 0.32 & -- & -- & -- & -- & -- \\
 & Judge-opt.\ vs direct, GPT & 0.26 & 0.69 & 0.38 & -- & -- & -- & -- & -- \\
HPO & Judge-opt.\ vs direct, Sonnet & 0.43 & 0.81 & 0.53 & -- & -- & -- & -- & -- \\
 & Judge-opt.\ vs direct, Opus & 0.42 & 0.90 & 0.47 & -- & -- & -- & -- & -- \\
 & Judge-opt.\ vs direct, GPT & 0.35 & 0.90 & 0.39 & -- & -- & -- & -- & -- \\
\bottomrule
\end{tabular}
\caption{The precision bound at other values (post hoc). Ratio: precision of the first arm over the comparison arm (mean of per-run precision); a column shows whether the bound is met at that multiple. The pre-registered bound is 0.8.}
\label{tab:bound}
\end{table}

%% file: appendix_round6_tables.tex
\begin{table}[h]
\centering\scriptsize
\setlength{\tabcolsep}{2.5pt}
\begin{tabular}{lllcccccc}
\toprule
Truth & Panel & Comparison & $\Delta R$ & runs only & crossed & $p_\times$ & credit, crossed ($p_\times$) & cost avoided, crossed ($p_\times$) \\
\midrule
$\tau=2$, all (main) & Claude LOO & vs direct & $+1.77$ & [$+1.31$, $+2.25$] & [$+0.38$, $+3.30$] & .015 & $+1.04$ [$-0.03$, $+2.13$] (.054) & $+0.73$ [$-0.09$, $+1.79$] (.086) \\
 &  & vs new direct & $+1.95$ & [$+1.45$, $+2.46$] & [$+0.67$, $+3.38$] & .004 & $+1.13$ [$+0.18$, $+2.15$] (.028) & $+0.82$ [$+0.03$, $+1.86$] (.039) \\
 &  & vs ablation & $+1.96$ & [$+1.56$, $+2.37$] & [$+0.36$, $+3.73$] & .024 & $+1.22$ [$+0.04$, $+2.37$] (.045) & $+0.74$ [$-0.23$, $+2.04$] (.137) \\
 & Claude+GPT LOO & vs direct & $+1.78$ & [$+1.33$, $+2.23$] & [$+0.55$, $+3.18$] & .009 & $+1.05$ [$+0.22$, $+1.93$] (.020) & $+0.72$ [$-0.15$, $+1.88$] (.102) \\
 &  & vs new direct & $+1.85$ & [$+1.39$, $+2.33$] & [$+0.72$, $+3.22$] & .002 & $+1.04$ [$+0.28$, $+1.87$] (.013) & $+0.81$ [$-0.00$, $+1.86$] (.051) \\
 &  & vs ablation & $+2.15$ & [$+1.75$, $+2.55$] & [$+0.68$, $+3.76$] & .004 & $+1.36$ [$+0.45$, $+2.33$] (.005) & $+0.79$ [$-0.29$, $+2.19$] (.145) \\
 & Sonnet own & vs direct & $+1.25$ & [$+0.72$, $+1.78$] & [$-0.49$, $+2.93$] & .116 & $+0.80$ [$-0.79$, $+2.33$] (.189) & $+0.45$ [$-0.11$, $+1.20$] (.117) \\
 &  & vs new direct & $+1.55$ & [$+1.05$, $+2.06$] & [$-0.10$, $+3.17$] & .062 & $+1.09$ [$-0.35$, $+2.50$] (.107) & $+0.46$ [$-0.08$, $+1.22$] (.102) \\
 &  & vs ablation & $+1.16$ & [$+0.76$, $+1.57$] & [$-1.07$, $+3.35$] & .192 & $+0.66$ [$-1.51$, $+2.64$] (.304) & $+0.50$ [$-0.21$, $+1.38$] (.165) \\
\midrule
$\tau=1$, all & Claude LOO & vs direct & $+1.08$ & [$+0.64$, $+1.53$] & [$+0.22$, $+2.01$] & .018 & $+1.05$ [$+0.35$, $+1.81$] (.006) & $+0.03$ [$-0.44$, $+0.57$] (.478) \\
 &  & vs new direct & $+1.44$ & [$+0.91$, $+1.97$] & [$+0.53$, $+2.40$] & .004 & $+1.02$ [$+0.37$, $+1.74$] (.003) & $+0.42$ [$-0.15$, $+1.07$] (.115) \\
 &  & vs ablation & $+0.69$ & [$+0.30$, $+1.09$] & [$-0.22$, $+1.61$] & .102 & $+1.12$ [$+0.35$, $+1.93$] (.007) & $-0.42$ [$-0.84$, $-0.08$] (.978) \\
$\tau=1.5$, all & Claude LOO & vs direct & $+1.14$ & [$+0.68$, $+1.59$] & [$+0.23$, $+2.13$] & .017 & $+1.09$ [$+0.25$, $+1.96$] (.013) & $+0.05$ [$-0.39$, $+0.58$] (.433) \\
 &  & vs new direct & $+1.51$ & [$+1.00$, $+2.07$] & [$+0.56$, $+2.53$] & .003 & $+1.15$ [$+0.37$, $+1.97$] (.004) & $+0.36$ [$-0.12$, $+0.92$] (.127) \\
 &  & vs ablation & $+1.12$ & [$+0.69$, $+1.52$] & [$+0.08$, $+2.15$] & .039 & $+1.29$ [$+0.41$, $+2.27$] (.008) & $-0.17$ [$-0.58$, $+0.29$] (.734) \\
$\tau=3$, all & Claude LOO & vs direct & $+1.59$ & [$+1.02$, $+2.18$] & [$+0.01$, $+3.24$] & .049 & $+0.78$ [$-0.44$, $+2.04$] (.143) & $+0.81$ [$-0.12$, $+1.91$] (.080) \\
 &  & vs new direct & $+1.84$ & [$+1.26$, $+2.44$] & [$+0.29$, $+3.42$] & .024 & $+0.94$ [$-0.19$, $+2.14$] (.087) & $+0.91$ [$-0.01$, $+2.00$] (.052) \\
 &  & vs ablation & $+1.91$ & [$+1.31$, $+2.50$] & [$+0.03$, $+3.79$] & .047 & $+0.95$ [$-0.49$, $+2.32$] (.134) & $+0.95$ [$-0.25$, $+2.34$] (.101) \\
 & Sonnet own & vs direct & $+1.15$ & [$+0.58$, $+1.72$] & [$-0.46$, $+2.79$] & .113 & $+0.76$ [$-0.73$, $+2.23$] (.189) & $+0.39$ [$-0.12$, $+1.07$] (.135) \\
 &  & vs new direct & $+1.50$ & [$+0.94$, $+2.07$] & [$-0.02$, $+3.05$] & .053 & $+1.10$ [$-0.28$, $+2.53$] (.098) & $+0.40$ [$-0.12$, $+1.14$] (.145) \\
 &  & vs ablation & $+1.16$ & [$+0.63$, $+1.70$] & [$-0.77$, $+3.06$] & .161 & $+0.65$ [$-1.26$, $+2.43$] (.285) & $+0.52$ [$-0.16$, $+1.39$] (.136) \\
$\tau=2$, 2 of 3 & Claude LOO & vs direct & $+1.63$ & [$+1.11$, $+2.13$] & [$+0.54$, $+2.79$] & .005 & $+1.19$ [$+0.36$, $+2.07$] (.008) & $+0.44$ [$-0.12$, $+1.22$] (.117) \\
 &  & vs new direct & $+1.78$ & [$+1.25$, $+2.33$] & [$+0.80$, $+2.84$] & .002 & $+1.21$ [$+0.41$, $+2.10$] (.006) & $+0.58$ [$+0.09$, $+1.18$] (.025) \\
 &  & vs ablation & $+1.73$ & [$+1.28$, $+2.17$] & [$+0.55$, $+3.01$] & .007 & $+1.39$ [$+0.45$, $+2.37$] (.006) & $+0.34$ [$-0.30$, $+1.24$] (.260) \\
$\tau=2$, mean & Claude LOO & vs direct & $+1.36$ & [$+0.86$, $+1.84$] & [$+0.36$, $+2.39$] & .011 & $+1.20$ [$+0.37$, $+2.09$] (.009) & $+0.16$ [$-0.24$, $+0.64$] (.288) \\
 &  & vs new direct & $+1.68$ & [$+1.13$, $+2.23$] & [$+0.70$, $+2.71$] & .003 & $+1.21$ [$+0.41$, $+2.08$] (.005) & $+0.46$ [$+0.03$, $+0.96$] (.036) \\
 &  & vs ablation & $+1.35$ & [$+0.93$, $+1.77$] & [$+0.27$, $+2.48$] & .017 & $+1.40$ [$+0.47$, $+2.39$] (.008) & $-0.05$ [$-0.40$, $+0.35$] (.612) \\
\bottomrule
\end{tabular}
\caption{Uncertainty of the stage-3 harness's gain when the recurring propositions are treated as a sample too (post hoc; Sonnet, NB201, 24 runs per arm). Runs only: 90\% interval from resampling runs within each arm, computed from this table's own draws, so it can differ from Table~\ref{tab:sel} in the second decimal. Crossed: runs are resampled within each arm and, jointly for both arms, the distinct propositions of the two arms are resampled with replacement (each run's $R$ is recomputed with every proposition weighted by its resampled multiplicity); 4,000 draws. $p_\times$: share of crossed draws $\leq 0$. Credit: $\sum_{\text{true}}(1-\pi)$; cost avoided: the comparison arm's $\sum_{\text{false}}\pi$ minus the harness's. The calibration is refitted under each truth definition and panel; at $\tau=3$ with a damped fit (Appendix~\ref{app:sens}).}
\label{tab:crossed}
\end{table}

%% file: appendix_panel.tex
\section{Panel dependence of the measurement results in detail (post hoc)}
\label{app:panel}
This appendix gives the post hoc panel analyses summarized in Section~\ref{sec:n}. All of it is hypothesis-generating: the panels share members, the AUC boundary below was read off the data, and panel quality is confounded with the presence of the other frontier model.

\paragraph{Panel strength and panel membership.}
Leaving the generator out of its own panel gave the two Claude generators panels of very different quality. Sonnet was scored by Haiku and Opus, whose prior predicts truth on NAS-Bench-201 with AUC .87, well above Sonnet's own .71; Opus was scored by Haiku and Sonnet (.74), well below its own .88. This matters because, relative to a panel, a true claim can be surprising for two reasons: the generator knew something the panel did not, or it learned it from evidence. Direct generation can earn $R$ only in the first way.

Post hoc, we therefore rescored each generator under every panel drawn from the other three models, seven per generator and benchmark (Figure~\ref{fig:panel}; Table~\ref{tab:panelfull}). Two properties of $R$ shape how these results should be read. First, as a panel approaches truth, every arm's $R$ shrinks toward zero by construction. Direct generation's $R$ does fall as the panel's AUC rises, in all six generator--benchmark combinations (Spearman $\rho$ from $-.68$ to $-.89$); for Opus on NAS-Bench-201 it falls from $+5.15$ per run under Haiku to $+1.03$ under GPT. Second, differences between arms shrink with it, so their correlation with AUC carries little information: Sonnet's gain on NAS-Bench-201 is negative under all seven panels, yet $\rho=+.86$. What matters is the sign of the difference. Below, a gain means one-sided $p<.05$, uncorrected.

On NAS-Bench-201, measurement raises $R$ significantly for Opus and GPT under all 8 panels that include the other frontier model (AUC $\geq .87$; $+0.84$ to $+1.71$, $p\leq.015$) and under none of the 6 that do not (AUC $\leq .74$; $-0.56$ to $+0.30$). No panel falls between these values. Among these panels, therefore, panel quality cannot be separated from the presence of the other frontier model, and the split is one contrast per generator, not 14 independent ones.

A generator's own blind ratings partly separate the two explanations: for Opus and GPT they form a strong panel (AUC .88--.95) that contains no other frontier model. They capture the generator's knowledge only in part, since direct generation still earns positive $R$ against them in five of six generator--benchmark pairs (Sonnet $+3.69$ and $+2.31$, Opus $+1.93$ and $+0.85$, GPT $+0.80$ on HPOBench; GPT on NAS-Bench-201 $-1.34$). Against these ratings (Table~\ref{tab:self}), measurement raises $R$ for Opus ($+0.91$ on NAS-Bench-201, $+0.48$ on HPOBench) and for GPT ($+2.69$, $+0.61$; Holm $p<.025$ over six pairs). This favors panel quality over panel membership as the explanation. With a prior-only calibration, the Opus gain on NAS-Bench-201 weakens to $+0.67$ ($p=.03$, Holm .09). Sonnet gains under no panel, including its own (AUC .71--.72). On HPOBench the panel matters little: GPT gains under all seven panels ($+0.42$ to $+1.01$), Opus under four and Sonnet under none. Because the panels share members and the boundary was read off the data, this analysis explains the pre-registered results; it does not test a new prediction.

\paragraph{Panel size or panel composition?}
Leave-one-out panels have two members for Claude generators and three for GPT, and averaging more members changes the spread of the mean prior. Grouping every panel by size (post hoc; Table~\ref{tab:sens_panel}) shows that size does not drive the measurement results. Among two-member Claude panels scoring GPT on NAS-Bench-201, Haiku and Sonnet give $+0.30$ ($p=.23$), while Haiku and Opus give $+1.43$ and Sonnet and Opus $+1.71$ ($p<.001$); Opus alone gives $+1.53$. For Opus, Haiku and Sonnet give $+0.19$, while GPT alone gives $+1.25$ and either Claude model with GPT $+0.84$ to $+0.91$. The stage-3 harness gains under every panel of every size ($+1.26$ to $+1.82$). For Opus on NAS-Bench-201, the measurement gain is also calibration-dependent: under the panel with GPT and under Opus's own ratings it is significant in 11 of 14 calibrations but not with a ridge penalty of 10 or 100 or a calibration fitted on the arms' own claims ($+0.56$, $-0.03$ and $+0.03$ with GPT in the panel, from the released sensitivity log; $+0.58$, $+0.04$ and $+0.02$ under Opus's own ratings, Table~\ref{tab:sens_cal_own}).

\paragraph{Evidence or tokens?}
Measurement costs several times the tokens of direct generation (Table~\ref{tab:tokens}), so a gain could reflect the extra tokens rather than the evidence. For Opus on NAS-Bench-201, the arms of stages 6--8 separate the two. They reuse the same 100 designed probes and repeat only the proposal call. That call costs about as many tokens with shuffled values (18.2k) as with exact values (17.6k), and fewer with values withheld (14.8k); the design call adds 15.1k to every arm. The pre-registered stage-7 test of these controls, under the original metric, was inconclusive. Post hoc, under $R$, exact values beat withheld values by 0.92 per run and shuffled values by 1.84 ($p=.005$ and $p<.001$). This holds even under the Claude panel, where no variant beats direct generation (largest difference $+0.19$), and the margins are larger under the other panels. Under every panel that predicts truth about as well as Opus, $R$ falls as the evidence gets worse (Table~\ref{tab:controls}): exact values gain most over direct generation ($+0.90$ to $+1.25$, $p\leq.009$), simulated noise and the 12-epoch proxy gain less, and withheld values, shuffled values and the design-only harness lose ($-0.30$ to $-1.02$). For Opus on NAS-Bench-201, then, the same probes with uninformative values, at similar or lower cost, do not produce the gain.

\input{tab_controls}

%% file: tab_controls.tex
\begin{table}[t]
\centering\small
\setlength{\tabcolsep}{4pt}
\begin{tabular}{lrrrrrrr}
\toprule
Panel & Exact & Noise & 2$\times$ noise & 12-ep.\ proxy & Withheld & Shuffled & Design only \\
\midrule
Claude leave-one-out (H+S) & $+$0.19 & $-$0.14 & $-$0.27 & $-$0.71 & $-$0.73 & $-$1.65 & $-$1.23 \\
H+S+G & $+$0.90$^{*}$ & $+$0.50 & $+$0.45 & $+$0.24 & $-$0.46 & $-$0.76 & $-$0.95 \\
G only & $+$1.25$^{*}$ & $+$0.76$^{*}$ & $+$0.72$^{*}$ & $+$0.65$^{*}$ & $-$0.30 & $-$0.63 & $-$0.77 \\
Opus's own ratings & $+$0.91$^{*}$ & $+$0.48 & $+$0.26 & $-$0.27 & $-$0.30 & $-$0.86 & $-$1.02 \\
\midrule
Precision (direct .87) & .75 & .71 & .69 & .65 & .79 & .58 & .69 \\
\bottomrule
\end{tabular}
\caption{Evidence or tokens? Opus on NB201, measurement variants and controls minus direct generation on $R$ (post hoc; 24 runs per arm; $^{*}p<.05$, one-sided, uncorrected). All arms but design-only reuse the same 100 designed probes and differ only in what the proposal prompt shows: exact values, values with simulated single-seed noise or doubled noise, values from a 12-epoch proxy, no values, or values shuffled across probes. Design-only writes 100 experiments it knows will not be run. H/S/G: Haiku/Sonnet/GPT.}
\label{tab:controls}
\end{table}

%% file: appendix_checks.tex
\section{Further checks (post hoc)}
\label{app:checks}
The checks in this appendix use cached runs, priors and tables only (no model calls). They are post hoc.

\paragraph{What $R$ sums over, and $R$ per distinct claim.}
$R$, precision and \SV$_3$ are computed over a run's distinct parseable claims: duplicates within a run are counted once, and on HPOBench the two wordings of the same proposition (Section~\ref{sec:setup}) are merged first. On NAS-Bench-201 nearly every run holds 10 distinct claims, so $R$ and $R$ per claim tell the same story. On HPOBench runs hold fewer, and the number differs between arms, which can move a sum such as $R$ toward zero for an arm with more duplicates. Table~\ref{tab:perclaim} therefore repeats the main comparisons with $R/n$, where $n$ is the run's number of distinct claims. The stage-3 harness's gains are unchanged (all three $p<.001$, under the leave-one-out panel and under Sonnet's own ratings). GPT's measurement gains on NAS-Bench-201 are unchanged as well. On HPOBench two contrasts weaken. Against the pooled direct baseline with the stage-3 rule, the gain in $R$ ($+0.94$, Table~\ref{tab:gpt}) is not significant per claim ($+0.055$, $p=.078$; under GPT's own ratings $+0.015$, $p=.34$), because that arm holds only 5.5 distinct claims per run against 7.8 for measurement. Against the 40-claim call, the pre-registered null is also null per claim ($-0.005$, $p=.44$). The judge-optimized harness's per-claim results have the same signs as its $R$ results.

\paragraph{The judge-optimized harness within claim type.}
The judge-optimized harness avoids the two claim types that are most often true, so part of its precision loss follows from base rates. Table~\ref{tab:judgetype} separates the two. On HPOBench, every claim it proposes is of the Other type, and its precision within that type (.35--.43) is far below direct generation's Other-type precision (.80--.96; 90\% bootstrap intervals of the difference $[-0.42, -0.32]$, $[-0.58, -0.49]$ and $[-0.55, -0.39]$ for Sonnet, Opus and GPT). The type mix alone would predict a precision of .80--.96, so on HPOBench the drop is not a type-shift artifact: the harness's claims are worse within type. On NAS-Bench-201 the two cannot be separated. There direct generation proposes almost no Other-type claims (8, 6 and 0 over 24 runs), so its within-type precision for the type that makes up 74--89\% of the judge-optimized harness's claims is not estimated. Within the one type both arms use often, claims that set another edge to \texttt{none}, the judge-optimized harness is less precise for Sonnet and Opus (.49 against .65; .52 against .89) and similar for GPT (.62 against .66).

\paragraph{Probe values and truth on NAS-Bench-201.}
On NAS-Bench-201 the probes report test accuracy and truth uses validation accuracy. The two splits give almost the same numbers. Over all 375,000 (parent, edit) pairs, the change in test accuracy correlates with the change in validation accuracy at $r=.999$, .999 and .997 (CIFAR-10, CIFAR-100, ImageNet16-120). Their signs agree in 98.5--98.9\% of pairs whose validation change is at least 0.5 pp, and they differ by 0.33--0.45 pp on average. On the probes the generators designed, $r$ is .999 (Sonnet), 1.000 (Opus) and .997 (GPT). Truth computed on the test split would label 99.8\% of the 1,740 claims the same way (198 true claims under each split, 196 in common). The separate split therefore gives almost no independence: on NAS-Bench-201, a probe is close to an exact sample of the quantity that truth averages, and a measure-first arm can in principle read truth off 100 well-chosen probes. The HPOBench probes are noisier (a single training seed, sign agreement 70\% with the 5-seed truth when the true difference is at least 0.5 pp; Section~\ref{sec:setup}).

\paragraph{Model versions.}
Claude models were called as \texttt{claude-haiku-5-5}, \texttt{claude-sonnet-5-5} and \texttt{claude-opus-5-5} through the Claude Code CLI, each call in its own temporary directory. The CLI's session logs record the model that answered each call. Every logged session on every study date reports exactly the requested identifier and no other (Table~\ref{tab:versions}). The counts are sessions logged from temporary directories on each date. They include calls from other work (all of 2026-10-20, see the pre-registration paragraph below) and exclude failed calls without a reply, so they do not match the paper's call counts exactly. GPT was called with the explicit model identifier \texttt{gpt-6-astra} (codex-cli 0.162.0, \texttt{codex exec --ephemeral}); ephemeral sessions keep no log, so we cannot audit the GPT version per call beyond the identifier we requested.

\begin{table}[h]
\centering\small
\begin{tabular}{lrrr}
\toprule
Date (UTC) & \texttt{claude-haiku-5-5} & \texttt{claude-sonnet-5-5} & \texttt{claude-opus-5-5} \\
\midrule
2026-10-20 & 766 & 384 & 240 \\
2026-10-21 & 455 & 1,636 & 1,150 \\
2026-10-22 & 35 & 34 & 110 \\
2026-10-23 & -- & 96 & 576 \\
2026-10-24 & -- & 24 & -- \\
\bottomrule
\end{tabular}
\caption{Claude Code CLI sessions by date and the model identifier each session's log reports (temporary-directory sessions; includes calls from other work, among them all 2026-10-20 sessions, which predate this study's first pre-registration and are pilots on other claim spaces). No session reports a different identifier.}
\label{tab:versions}
\end{table}

\paragraph{Which pre-registered criteria include the precision bound.}
The pre-registration documents are written in Chinese; we translate the decision criteria. Stage 10 (GPT, measurement against direct generation, each benchmark): ``M: supported: A-100's \SV{} is above BASE, one-sided permutation test, Holm-corrected together with A-SV, $p<.05$, and precision $\geq 0.8\times$ BASE; refuted: the upper bound of the difference's 90\% CI $<0.5$; otherwise inconclusive.'' Stages 2, 4, 5, 6, 8, 9 and 12 state the same precision condition for their measurement or harness tests. Stage 13b (the stage-3 harness against direct generation): ``supported: $p<.05$, and precision $\geq 0.8\times$ BASE; refuted: the upper bound of the 90\% CI of $\Delta R$ $<0.5$; otherwise inconclusive.'' Stage 13c (GPT, measurement against pooled direct generation): ``supported (the gain from measurement is not only more tokens): Holm $p<.05$; refuted (tokens suffice to explain it): the upper bound of the 90\% CI of A $-$ D4 $<0.5$; otherwise inconclusive.'' Stage 14a (pooled with the stage-3 rule) and stage 15c (the 40-claim call) use criteria ``of the same form as 13c''; stage 14b (the likely-true ablation), stage 15a (new direct runs) and stage 15b (shuffled and withheld values) state $p<.05$ (Holm-corrected where there are two tests) and the same refutation rule. None of stages 13c, 14 or 15 includes a precision condition, so their verdicts in Table~\ref{tab:ledger} follow from $p$ alone and are as pre-registered. Applying the bound to them is our own post hoc reading: for GPT on NAS-Bench-201, measure-first precision (.50) is .61 to .72 times that of every same-cost arm (pooled direct .70, pooled with the stage-3 rule .73, 40-claim call .82, withheld values .69), so those gains are supported as pre-registered but do not count as improvements under the bound (Table~\ref{tab:gpt}). On HPOBench the ratios are .97 to 1.63. The stage-3 harness's precision is 1.05, 1.12 and 1.00 times that of its three comparison arms.

\paragraph{A calibration with all three claim types.}
The default calibration separates self-conditions from all other claims but not the two other types, claims whose condition removes an edge and the remaining Other claims, whose base rates differ (.173 and .067 on NAS-Bench-201). The +Type variant of Table~\ref{tab:sens_cal} adds operation indicators, which are not these types. We therefore added an indicator for removes-an-edge claims and its interaction with the prior's logit to the default covariates (post hoc; NAS-Bench-201 only, since HPOBench has no such type). The stage-3 harness's gains barely move: $+1.73$, $+1.90$ and $+1.89$ against direct generation, the new stage-15 direct runs and the ablation (default $+1.77$, $+1.95$, $+1.96$; all $p<.001$), and $+1.24$, $+1.54$ and $+1.16$ under Sonnet's own ratings. GPT's NAS-Bench-201 measurement gains change by at most 0.22 and stay significant (against direct generation $+1.39$; pooled $+2.05$; pooled with the rule $+1.13$; 40-claim call $+0.86$; shuffled $+1.45$; withheld $+0.67$, $p=.021$). Opus's gain under the Claude panel stays null ($+0.18$). The judge-optimized harness's rise in $R$ for GPT on NAS-Bench-201 shrinks from $+1.04$ to $+0.79$.

\paragraph{GPT: same-cost baselines in detail.}
A pooled direct baseline ran direct generation four times and asked GPT to keep the 10 pooled claims most likely to be true, at about the cost of measurement. Measurement beat it on both benchmarks. Pooling and selecting changed $R$ by $-0.64$ and $+0.24$ relative to a single direct run, at or above direct precision (.70 against .69, .96 against .90). Applying the stage-3 harness's selection rule to the same pools did not close the gap either. Four pooled direct runs are redundant, however: their 40 hypotheses hold only 21.3 distinct claims on NAS-Bench-201 and 14.0 on HPOBench (post hoc). A single call that writes 40 claims and keeps the 10 most likely to be true is less redundant (36.8 and 25.3 distinct claims) and raises $R$ over a single direct run ($+0.73$, $p=.014$; $+0.85$, $p<.001$) at about 11k tokens per run. On NAS-Bench-201 measurement still beat it. On HPOBench it did not: there a single 40-claim call matched measurement at about two thirds of its tokens (11.0k against 17.4k), with precision .97 against .95. The same-budget advantage of measurement for GPT therefore holds on NAS-Bench-201 but not on HPOBench. Post hoc, against GPT's own blind ratings, the pooled baseline gained nothing over direct generation ($-0.28$, $-0.10$) while measurement did ($+2.69$, $+0.61$). On NAS-Bench-201 the 96 pooled direct runs contain no claims of the non-structural ``Other'' type (0 of 954), which make up 60\% of GPT's measure-first hypotheses, so no selection from these pools can reach them.
On NAS-Bench-201, GPT's direct claims hold as often as the calibrated Claude panel expects (.69 against .69), while its measure-first claims were expected to hold at .33 and hold at .50, mostly non-structural interactions that the panel rates unlikely. Per distinct claim (Table~\ref{tab:perclaim}), every NAS-Bench-201 contrast of Table~\ref{tab:gpt} keeps $p\leq.005$, and so do the HPOBench contrasts with direct generation, pooled direct generation, and shuffled and withheld values; the refuted contrast with the 40-claim call stays null ($-0.005$).

\paragraph{Stages and pre-registration documents.}
The study has 17 pre-registered stages in 16 documents: stages 1 to 15 and stage 12b each have their own document, and stage 4b (the Opus confirmation) is an addendum to the stage-4 document, written before its calls. Table~\ref{tab:ledger} lists each stage's criteria and verdicts.
Table~\ref{tab:prereg} lists the documents with their SHA-256 hashes and file times. The documents themselves are in the supplementary material and at \url{https://osf.io/c2hkg/?view_only=d05d8b1acd504e6ab87a242522b769df}, so the hashes can be checked against the text we cite. The times come from the local filesystem and are not attested by a third party; a deposit made now cannot prove when a document was written, and the table is offered as a record, not as proof. Only one document changed after it was created: \texttt{PREREG\_s4.md}, 15 minutes later, when the stage-4b addendum was added. In the study's directory the only files older than the first document are the claim-space code and one literature-search cache entry. In the CLI session logs, the first session whose prompt poses a claim about either benchmark starts at 01:05 UTC on 2026-10-21, nine minutes after \texttt{PREREG\_s1.md}; all 1,504 earlier sessions in Table~\ref{tab:versions} (every 2026-10-20 session and 114 on 2026-10-21) predate that document. A keyword search finds neither benchmark in them, except two sessions that ask for a review of an unrelated manuscript whose text mentions NAS-Bench-201. They are exploratory pilots on other claim spaces (with their own pre-registrations) and unrelated work, and none is used as evidence here. Dates written inside the documents are in local time (UTC$-$7), up to a day behind the UTC times used here; for example, \texttt{PREREG\_s15.md} dates the stage-5 direct runs, made after 06:04 UTC on 2026-10-21, to ``10-20''.
\input{appendix_prereg_table}

\input{appendix_checks_tables}

%% file: appendix_prereg_table.tex
\begin{table}[h]
\centering\scriptsize
\setlength{\tabcolsep}{3pt}
\begin{tabular}{llll}
\toprule
Document & Created (UTC) & Last modified (UTC) & SHA-256 \\
\midrule
\texttt{PREREG\_s1.md} & 2026-10-21 00:56 & 2026-10-21 00:56 & \texttt{9d7125b6b956db218b5252ec4a8eda318ef03de5d72a94c941e89ec0aaa57112} \\
\texttt{PREREG\_s2.md} & 2026-10-21 03:56 & 2026-10-21 03:56 & \texttt{ff825cd04b8ccb9a6c0d06804584b5ff21e384014d4e64d7af32010abe1de50c} \\
\texttt{PREREG\_s3.md} & 2026-10-21 04:53 & 2026-10-21 04:53 & \texttt{07f52c56b97a846b20fa78be1e0fd31159aae5197448627d0e4fd3cd32aa005a} \\
\texttt{PREREG\_s4.md} & 2026-10-21 05:37 & 2026-10-21 05:52 & \texttt{2a9c42cdb788d5c2c19c21657929bc4277b1065e97d3d1431db16b10c7551ee5} \\
\texttt{PREREG\_s5.md} & 2026-10-21 06:04 & 2026-10-21 06:04 & \texttt{52d567d9091c092c341e06b02dacbb3bf97bb55ab83936566b3fa88b2a448840} \\
\texttt{PREREG\_s6.md} & 2026-10-21 07:29 & 2026-10-21 07:29 & \texttt{2c2f7d9899d4210bbef90f91694a6d10ff27716c047862f458398dbfbde0f493} \\
\texttt{PREREG\_s7.md} & 2026-10-21 08:04 & 2026-10-21 08:04 & \texttt{7013979422c869a0dc159786fdec1fed31ad0cea00c3e50edbf09b8e3f32a569} \\
\texttt{PREREG\_s8.md} & 2026-10-21 08:15 & 2026-10-21 08:15 & \texttt{8df7d217a437e8a56b7c1a735b5da29f65f7697258e844fd735bc0279d939643} \\
\texttt{PREREG\_s9.md} & 2026-10-21 08:44 & 2026-10-21 08:44 & \texttt{99ab055f56cecab5fec9c2c4346579897e1b5491f05f393608b05dc1495feec4} \\
\texttt{PREREG\_s10.md} & 2026-10-21 18:58 & 2026-10-21 18:58 & \texttt{9cf1f532e70bd80fb001f9fcd80a415d2f29f9827608c3c9dcb5c73f1916665c} \\
\texttt{PREREG\_s11.md} & 2026-10-21 23:31 & 2026-10-21 23:31 & \texttt{7bad9c7e24822fdbf6190eacacf4cae49a24f5522c1b4d532e30f39ee31b4bb2} \\
\texttt{PREREG\_s12.md} & 2026-10-22 03:43 & 2026-10-22 03:43 & \texttt{eeed4897748232b2eb411e81b552e21694c5ea022b574b63a79ea75db4a02671} \\
\texttt{PREREG\_s12b.md} & 2026-10-22 05:37 & 2026-10-22 05:37 & \texttt{158e59b535e715ee245d12b6c95c81162e8208d4114f91912eaee107312075dd} \\
\texttt{PREREG\_s13.md} & 2026-10-23 05:21 & 2026-10-23 05:21 & \texttt{1db098e8f4aba34a33daf086a80157d3529a5c8e8e13f015671b896db7c79aa7} \\
\texttt{PREREG\_s14.md} & 2026-10-23 18:41 & 2026-10-23 18:41 & \texttt{c37ca970ea6c17e170f25fc9c99c0d635f156858a597c888a3a5821f167e2857} \\
\texttt{PREREG\_s15.md} & 2026-10-24 01:42 & 2026-10-24 01:42 & \texttt{9a4baae40153af6de60f80eb0064aa0488b5c2def43946b845cd4bafa86a5d05} \\
\bottomrule
\end{tabular}
\caption{The 16 stage pre-registration documents (stage 4b is an addendum inside \texttt{PREREG\_s4.md}). Times are the local filesystem's creation and last-modification times, converted to UTC; they are not third-party timestamps. The documents are in the supplementary material and in an anonymized OSF deposit, \url{https://osf.io/c2hkg/?view_only=d05d8b1acd504e6ab87a242522b769df}.}
\label{tab:prereg}
\end{table}

%% file: appendix_checks_tables.tex
\begin{table}[h]
\centering\scriptsize
\setlength{\tabcolsep}{3pt}
\begin{tabular}{llcccccc}
\toprule
 & & \multicolumn{2}{c}{Distinct claims per run} & \multicolumn{2}{c}{Leave-one-out Claude panel} & \multicolumn{2}{c}{Generator's own ratings} \\
\cmidrule(lr){3-4}\cmidrule(lr){5-6}\cmidrule(lr){7-8}
Bench. & Comparison & First arm & Second arm & $\Delta R$ & $\Delta(R/n)$ ($p$) & $\Delta R$ & $\Delta(R/n)$ ($p$) \\
\midrule
NB201 & Stage-3 harness vs direct & 10.00 & 10.00 & $+1.77$ & $+0.177$ ($<.001$) & $+1.25$ & $+0.125$ ($<.001$) \\
 & \quad vs new direct & 10.00 & 10.00 & $+1.95$ & $+0.195$ ($<.001$) & $+1.55$ & $+0.155$ ($<.001$) \\
 & \quad vs likely-true ablation & 10.00 & 9.96 & $+1.96$ & $+0.196$ ($<.001$) & $+1.16$ & $+0.114$ ($<.001$) \\
 & GPT measure-first vs direct & 10.00 & 10.00 & $+1.60$ & $+0.160$ ($<.001$) & $+2.69$ & $+0.269$ ($<.001$) \\
 & \quad vs pooled direct & 10.00 & 10.00 & $+2.24$ & $+0.224$ ($<.001$) & $+2.97$ & $+0.297$ ($<.001$) \\
 & \quad vs pooled + stage-3 rule & 10.00 & 10.00 & $+1.16$ & $+0.116$ ($<.001$) & $+2.02$ & $+0.202$ ($<.001$) \\
 & \quad vs 40-claim call & 10.00 & 9.79 & $+0.88$ & $+0.086$ (.005) & $+1.28$ & $+0.128$ ($<.001$) \\
 & \quad vs shuffled values & 10.00 & 10.00 & $+1.40$ & $+0.140$ ($<.001$) & $+0.98$ & $+0.098$ ($<.001$) \\
 & \quad vs withheld values & 10.00 & 10.00 & $+0.89$ & $+0.089$ (.004) & $+1.89$ & $+0.189$ ($<.001$) \\
 & Opus measure-first vs direct & 10.00 & 9.88 & $+0.19$ & $+0.013$ (.366) & $+0.91$ & $+0.088$ (.009) \\
HPO & GPT measure-first vs direct & 7.79 & 7.96 & $+0.94$ & $+0.119$ ($<.001$) & $+0.61$ & $+0.076$ (.002) \\
 & \quad vs pooled direct & 7.79 & 7.46 & $+0.70$ & $+0.081$ ($<.001$) & $+0.72$ & $+0.087$ ($<.001$) \\
 & \quad vs pooled + stage-3 rule & 7.79 & 5.50 & $+0.94$ & $+0.055$ (.078) & $+0.53$ & $+0.015$ (.335) \\
 & \quad vs 40-claim call & 7.79 & 7.33 & $+0.09$ & $-0.005$ (.444) & $+0.42$ & $+0.045$ (.032) \\
 & \quad vs shuffled values & 7.79 & 8.54 & $+1.42$ & $+0.181$ ($<.001$) & $+1.01$ & $+0.128$ ($<.001$) \\
 & \quad vs withheld values & 7.79 & 8.96 & $+0.85$ & $+0.127$ ($<.001$) & $+0.51$ & $+0.078$ (.003) \\
 & Opus measure-first vs direct & 9.67 & 9.54 & $+0.42$ & $+0.041$ (.056) & $+0.48$ & $+0.048$ (.004) \\
NB201 & Judge-opt.\ vs direct, Sonnet & 10.00 & 10.00 & $+0.88$ & $+0.088$ (.006) & $-1.05$ & $-0.105$ (.003) \\
 & Judge-opt.\ vs direct, Opus & 10.00 & 9.88 & $-2.24$ & $-0.230$ ($<.001$) & $-1.07$ & $-0.109$ ($<.001$) \\
 & Judge-opt.\ vs direct, GPT & 10.00 & 10.00 & $+1.04$ & $+0.104$ (.003) & $+1.91$ & $+0.191$ ($<.001$) \\
HPO & Judge-opt.\ vs direct, Sonnet & 10.00 & 9.38 & $-0.89$ & $-0.096$ ($<.001$) & $-1.63$ & $-0.179$ ($<.001$) \\
 & Judge-opt.\ vs direct, Opus & 10.00 & 9.54 & $-1.69$ & $-0.182$ ($<.001$) & $-1.24$ & $-0.127$ ($<.001$) \\
 & Judge-opt.\ vs direct, GPT & 10.00 & 7.96 & $-0.87$ & $-0.121$ ($<.001$) & $-0.58$ & $-0.082$ ($<.001$) \\
\bottomrule
\end{tabular}
\caption{$R$ per distinct claim (post hoc). $R$ sums over a run's distinct parseable claims, after duplicates and, on HPOBench, equivalent wordings are merged; $n$ is that number. $\Delta(R/n)$: difference in the per-run mean of $R/n$, with a one-sided permutation $p$ in the direction shown. Paper-default calibration.}
\label{tab:perclaim}
\end{table}

\begin{table}[h]
\centering\scriptsize
\setlength{\tabcolsep}{3pt}
\begin{tabular}{llccccccc}
\toprule
 & & \multicolumn{2}{c}{Precision} & & \multicolumn{4}{c}{Share of claims and precision within type: judge-opt.\ / direct} \\
\cmidrule(lr){3-4}\cmidrule(lr){6-9}
Bench. & Gen. & Judge-opt. & Direct & Type-adjusted & Type & Share & Precision & $n$ \\
\midrule
NB201 & Sonnet & .43 & .79 & .36 & Self-condition & .03 / .68 & .57 / .87 & 7 / 164 \\
 & & & & & Removes an edge & .23 / .28 & .49 / .65 & 55 / 68 \\
 & & & & & Other & .74 / .03 & .40 / .25 & 178 / 8 \\
NB201 & Opus & .28 & .86 & .71 & Self-condition & .00 / .31 & -- / .84 & 0 / 73 \\
 & & & & & Removes an edge & .17 / .67 & .52 / .89 & 42 / 158 \\
 & & & & & Other & .82 / .03 & .22 / .67 & 198 / 6 \\
NB201 & GPT & .26 & .69 & -- & Self-condition & .00 / .23 & -- / .79 & 0 / 56 \\
 & & & & & Removes an edge & .11 / .77 & .62 / .66 & 26 / 184 \\
 & & & & & Other & .89 / .00 & .22 / -- & 214 / 0 \\
HPO & Sonnet & .43 & .81 & .80 & Self-condition & .00 / .44 & -- / .82 & 0 / 98 \\
 & & & & & Other & 1.00 / .56 & .43 / .80 & 240 / 127 \\
HPO & Opus & .42 & .90 & .96 & Self-condition & .00 / .55 & -- / .85 & 0 / 127 \\
 & & & & & Other & 1.00 / .45 & .42 / .96 & 240 / 102 \\
HPO & GPT & .35 & .89 & .82 & Self-condition & .00 / .62 & -- / .93 & 0 / 118 \\
 & & & & & Other & 1.00 / .38 & .35 / .82 & 240 / 73 \\
\bottomrule
\end{tabular}
\caption{Precision of the judge-optimized harness within claim type (post hoc; claims pooled over 24 runs per arm, so precision can differ by .01 from the per-run means of Table~\ref{tab:a}). Type-adjusted: the precision implied by the judge-optimized harness's mix of claim types alone, its type shares times direct generation's within-type precision; ``--'' where direct generation proposed no claim of a type the harness used. On NB201, direct generation proposed only 8, 6 and 0 Other claims, so the type-adjusted values there rest on very few claims. Removes an edge: the condition sets another edge to \texttt{none} (NB201 only).}
\label{tab:judgetype}
\end{table}

%% file: appendix_tables.tex
\begin{table}[h]
\centering\small
\begin{tabular}{llcccc}
\toprule
Bench. & Gen. & Direct & Measure-first & Judge-opt. & Pooled direct \\
\midrule
NB201 & Sonnet & 5.2k & 30.9k & 33.3k & -- \\
 & Opus & 5.9k & 32.6k & 35.1k & -- \\
 & GPT & 3.5k & 21.0k & 25.2k & 21.2k \\
HPO & Sonnet & 5.2k & 27.3k & 32.2k & -- \\
 & Opus & 5.4k & 36.5k & 34.4k & -- \\
 & GPT & 3.6k & 17.4k & 24.2k & 20.4k \\
\bottomrule
\end{tabular}
\caption{Mean tokens per run, 24 runs per arm. Claude counts are the CLI-reported input, output and cache tokens, including the CLI's own system prompt; GPT counts are codex-cli's ``tokens used''. The two are not comparable with each other. Probes are table lookups and use no tokens or compute themselves; the measure-first cost is the design call plus the longer proposal prompt that lists the 100 results. The optimizer capped harnesses at 40k tokens per run. Pooled direct (stage 13, GPT only) runs direct generation four times and keeps ten of the pooled claims.}
\label{tab:tokens}
\end{table}

\begin{table}[h]
\centering\small
\begin{tabular}{lllrrrrrr}
\toprule
Gen. & Stage & Harness & $n$ & $\Delta$prec. & $\Delta$\SV$_3$ & Holm & $\Delta R$ & Holm \\
\midrule
Sonnet & 1 & L9 cell (0,1,1,1) & 8 & $-$.09 & $-$1.50 & 1 & $-$2.02 & 1 \\
 & 1 & L9 cell (0,2,2,2) & 8 & $-$.45 & $-$1.12 & 1 & $-$1.15 & 1 \\
 & 1 & L9 cell (1,0,1,2) & 8 & $-$.03 & $+$2.00 & .343 & $+$1.08 & .405 \\
 & 1 & L9 cell (1,1,2,0) & 8 & $-$.51 & $-$2.12 & 1 & $-$1.43 & 1 \\
 & 1 & L9 cell (1,2,0,1) & 8 & $-$.01 & $-$0.75 & 1 & $-$1.25 & 1 \\
 & 1 & L9 cell (2,0,2,1) & 8 & $+$.03 & $-$0.12 & 1 & $-$1.09 & 1 \\
 & 1 & L9 cell (2,1,0,2) & 8 & $-$.56 & $-$1.75 & 1 & $-$1.40 & 1 \\
 & 1 & L9 cell (2,2,1,0) & 8 & $-$.14 & $-$0.50 & 1 & $-$0.71 & 1 \\
 & 2 & judge-fitness final & 8 & $-$.49 & $-$0.62 & 1 & $-$0.09 & 1 \\
 & 2 & validity-fitness final & 8 & $+$.01 & $-$0.75 & 1 & $-$0.54 & 1 \\
 & 3 & \SV{}-fitness rerun final & 8 & $+$.14 & $+$2.00 & .172 & $+$1.98 & .023 \\
 & 5 & judge-optimized (24 runs) & 24 & $-$.36 & $+$0.71 & .385 & $+$0.88 & .038 \\
 & 8 & design-only, 100 probes & 24 & $-$.09 & $-$0.04 & 1 & $-$0.29 & 1 \\
 & 13 & \SV{}-fitness rerun, confirmation & 24 & $+$.04 & $+$1.63 & -- & $+$1.77 & -- \\
Opus & 3 & judge-fitness final & 8 & $-$.54 & $-$4.00 & 1 & $-$1.56 & 1 \\
 & 3 & validity-fitness final & 8 & $-$.06 & $-$0.75 & 1 & $-$0.58 & 1 \\
 & 3 & prior-avoiding selection & 8 & $-$.39 & $-$3.00 & 1 & $-$1.27 & 1 \\
 & 3 & cross-domain retrieval & 8 & $-$.14 & $-$0.62 & 1 & $+$0.04 & 1 \\
 & 5 & judge-optimized (24 runs) & 24 & $-$.59 & $-$4.21 & 1 & $-$2.24 & 1 \\
 & 8 & design-only, 100 probes & 24 & $-$.18 & $-$1.75 & 1 & $-$1.23 & 1 \\
\bottomrule
\end{tabular}
\caption{Every internal-reasoning harness evaluated on fresh NB201 runs, minus direct generation from the same stage, under the leave-one-out Claude panel (post hoc rescoring). L9 cells give the levels of retrieval, roles, challenge and selection; level 0 is direct generation. Holm correction is within each generator. Stage 13 is a pre-registered confirmation on 24 new seeds, outside the Holm family (one-sided $p<.001$ for $\Delta$\SV$_3$ and $\Delta R$).}
\label{tab:harness}
\end{table}

\begin{table}[h]
\centering\scriptsize
\begin{tabular}{llcccc@{\hskip 14pt}llcccc}
\toprule
Gen. & Panel & AUC & $R$ direct & $\Delta R$ & $p$ & Gen. & Panel & AUC & $R$ direct & $\Delta R$ & $p$ \\
\midrule
\multicolumn{12}{l}{\emph{NB201}} \\
Sonnet & H & .67 & $+$3.74 & $-$1.97 & 1 & Opus & H+S+G & .90 & $+$1.36 & $+$0.90 & .008 \\
Sonnet & H+O & .87 & $+$1.10 & $-$0.57 & .954 & Opus & G & .91 & $+$1.03 & $+$1.25 & $<$.001 \\
Sonnet & O & .88 & $+$1.61 & $-$0.90 & .995 & Opus & S+G & .91 & $+$1.46 & $+$0.84 & .015 \\
Sonnet & H+G & .90 & $+$0.72 & $-$0.30 & .823 & GPT & H & .67 & $+$3.55 & $-$0.56 & .909 \\
Sonnet & G & .91 & $+$0.57 & $-$0.17 & .693 & GPT & S & .71 & $+$3.25 & $-$0.12 & .613 \\
Sonnet & H+O+G & .92 & $+$0.40 & $-$0.10 & .630 & GPT & H+S & .74 & $+$2.30 & $+$0.30 & .222 \\
Sonnet & O+G & .92 & $+$0.61 & $-$0.23 & .762 & GPT & H+S+O & .87 & $+$0.03 & $+$1.60 & $<$.001 \\
Opus & H & .67 & $+$5.15 & $-$0.31 & .769 & GPT & H+O & .87 & $-$0.04 & $+$1.43 & $<$.001 \\
Opus & S & .71 & $+$4.59 & $-$0.21 & .696 & GPT & O & .88 & $-$0.02 & $+$1.53 & $<$.001 \\
Opus & H+S & .74 & $+$3.71 & $+$0.19 & .332 & GPT & S+O & .88 & $+$0.10 & $+$1.71 & $<$.001 \\
Opus & H+G & .90 & $+$1.50 & $+$0.91 & .007 & & & & & & \\
\multicolumn{12}{l}{\emph{HPO}} \\
Sonnet & H & .69 & $+$4.51 & $-$0.52 & .958 & Opus & S+G & .94 & $+$1.00 & $+$0.60 & .006 \\
Sonnet & H+O & .91 & $+$0.93 & $+$0.11 & .307 & Opus & H+G & .94 & $+$1.76 & $+$0.34 & .089 \\
Sonnet & O & .93 & $+$0.58 & $-$0.00 & .508 & Opus & G & .95 & $+$1.42 & $+$0.40 & .038 \\
Sonnet & H+G & .94 & $+$1.15 & $-$0.07 & .625 & GPT & H & .69 & $+$4.29 & $+$0.42 & .040 \\
Sonnet & G & .95 & $+$1.00 & $-$0.15 & .774 & GPT & S & .72 & $+$2.79 & $+$0.86 & $<$.001 \\
Sonnet & H+O+G & .95 & $+$0.38 & $+$0.05 & .399 & GPT & H+S & .74 & $+$2.69 & $+$0.74 & .001 \\
Sonnet & O+G & .96 & $+$0.23 & $+$0.07 & .341 & GPT & H+S+O & .91 & $+$1.18 & $+$0.94 & $<$.001 \\
Opus & H & .69 & $+$5.29 & $+$0.04 & .439 & GPT & S+O & .91 & $+$0.82 & $+$1.01 & $<$.001 \\
Opus & S & .72 & $+$2.89 & $+$0.54 & .015 & GPT & H+O & .91 & $+$1.28 & $+$0.72 & $<$.001 \\
Opus & H+S & .74 & $+$2.97 & $+$0.42 & .065 & GPT & O & .93 & $+$0.59 & $+$0.78 & $<$.001 \\
Opus & H+S+G & .92 & $+$1.28 & $+$0.52 & .017 & & & & & & \\
\bottomrule
\end{tabular}
\caption{Measure-first minus direct generation on $R$ under every panel drawn from the other three models (post hoc; 24 runs per arm; one-sided permutation $p$, uncorrected). H/S/O/G: Haiku/Sonnet/Opus/GPT.}
\label{tab:panelfull}
\end{table}

\begin{table}[h]
\centering\small
\begin{tabular}{llrrrrrr}
\toprule
 & & \multicolumn{3}{c}{Full calibration} & \multicolumn{3}{c}{Prior-only calibration} \\
\cmidrule(lr){3-5}\cmidrule(lr){6-8}
Bench. & Gen. & $\Delta R$ & $p$ & Holm & $\Delta R$ & $p$ & Holm \\
\midrule
NB201 & Sonnet & $-$2.05 & 1 & 1 & $-$1.90 & 1 & 1 \\
 & Opus & $+$0.91 & .007 & .021 & $+$0.67 & .029 & .088 \\
 & GPT & $+$2.69 & $<$.001 & $<$.001 & $+$2.46 & $<$.001 & $<$.001 \\
HPO & Sonnet & $+$0.25 & .115 & .230 & $-$0.30 & .897 & 1 \\
 & Opus & $+$0.48 & .004 & .017 & $+$0.47 & .009 & .038 \\
 & GPT & $+$0.61 & .002 & .008 & $+$0.61 & .001 & .007 \\
\bottomrule
\end{tabular}
\caption{Measure-first minus direct generation on $R$ when each generator is scored against its own blind three-way prior (post hoc; 24 runs per arm; one-sided permutation $p$; Holm over the six pairs).}
\label{tab:self}
\end{table}